\documentclass[lettersize,journal]{IEEEtran}

\usepackage{graphics}
\usepackage{graphicx}
\usepackage{epsfig}
\usepackage{multirow}
\usepackage{amssymb}
\usepackage{amsmath}
\usepackage{textcomp}
\usepackage{xspace}
\usepackage{threeparttable}
\usepackage{siunitx}

\usepackage{amsthm}
\usepackage{mathrsfs}

\usepackage{booktabs}
\usepackage{stackengine}
\usepackage{xfrac}
\usepackage{tabularx}
\usepackage{multirow}
\usepackage{cite}
\usepackage{lipsum}
\usepackage[caption=false]{subfig}
\usepackage{cancel}
\usepackage[normalem]{ulem} 
\usepackage{bm}
\usepackage{soul} 
\usepackage{stmaryrd} 

\usepackage[algo2e,linesnumbered,ruled,vlined]{algorithm2e}

\usepackage{enumitem}

\usepackage[integrals]{wasysym}

\usepackage{threeparttable}
\usepackage{optidef}

\makeatletter
\let\NAT@parse\undefined

\makeatother
\usepackage[hidelinks]{hyperref}
\hypersetup{
    colorlinks=true,
    linkcolor=blue,
    filecolor=magenta,
    urlcolor=cyan,
    citecolor=blue
}

\usepackage[dvipsnames]{xcolor,colortbl}
\definecolor{blockgray}{gray}{0.94}
\newtheorem{assumption}{Assumption}

\newtheorem{remark}{Remark}

\newcommand{\stdcell}[2]{%
\begin{tabular}[c]{@{}c@{}}
#1\\
$\pm$ #2
\end{tabular}}

\begin{document}

\title{Trajectory Prediction via Bayesian Intention Inference under Unknown Goals and Kinematics}

\author{Shunan Yin, Zehui Lu, Shaoshuai Mou
\thanks{This material is based upon work supported by the Office of Naval Research (ONR) and Saab, Inc. under the Threat and Situational Understanding of Networked Online Machine Intelligence (TSUNOMI) program (grant no. N00014-23-C-1016). Any opinions, ﬁndings and conclusions or recommendations expressed in this material are those of the author(s) and do not necessarily reﬂect the views of the ONR, the U.S. Government, or Saab, Inc.}
\thanks{S. Yin and S. Mou are with the School of Aeronautics and Astronautics, Purdue University, West Lafayette, IN 47907, USA {\tt\small \{yin249,mous\}@purdue.edu} }
\thanks{Z. Lu is an independent researcher, Fremont, CA 94538, USA {\tt\small zehuilu789@gmail.com}}
\thanks{The experimental video can be found in \url{https://youtu.be/7c48M3qHm64}}
}

\maketitle

\begin{abstract}
This work introduces an adaptive Bayesian algorithm for real-time trajectory prediction via intention inference, where a target’s intentions and motion characteristics are unknown and subject to change. The method concurrently estimates two critical variables: the target’s current intention, modeled as a Markovian latent state, and an intention parameter that describes the target’s adherence to a shortest-path policy. By integrating this joint update technique, the proposed algorithm maintains robustness against abrupt intention shifts in trajectory prediction and unknown motion dynamics. A sampling-based trajectory prediction mechanism then exploits these adaptive estimates to generate probabilistic forecasts with quantified uncertainty. We validate the algorithm through numerical experiments: Ablation studies of two cases, and a 500-trial Monte Carlo analysis; Hardware demonstrations on quadrotor and quadrupedal platforms. Experimental results demonstrate that the proposed approach significantly outperforms non-adaptive and partially adaptive methods. The method operates in real time around \SI{547}{Hz} without requiring training or detailed prior knowledge of target behavior, showcasing its applicability in various robotic systems.
\end{abstract}


\section{Introduction}
Real-world autonomous systems such as self-driving cars, service robots, and surveillance drones frequently face intention inference tasks~\cite{rudenko2020human}: they must determine what another agent or human is trying to achieve and where it is likely to go next~\cite{Lu2025FAPP, Kim2014Catching}. These tasks are inherently challenging for several reasons. First, the target’s motion dynamics are often unknown. For example, a pedestrian may switch between walking, jogging, or stopping unpredictably. Second, the agent’s intention may shift during execution, such as changing to a new goal without any observable signal, i.e., in a non-cooperative fashion. Finally, in many practical settings, only sparse or partial observations of the trajectory are available, and no labeled demonstrations can be assumed in advance. As a result, intention inference and trajectory prediction under such uncertainty have become active research areas, with numerous methods proposed to address these challenges.

\begin{figure}
\centering
\includegraphics[width=0.70\linewidth]{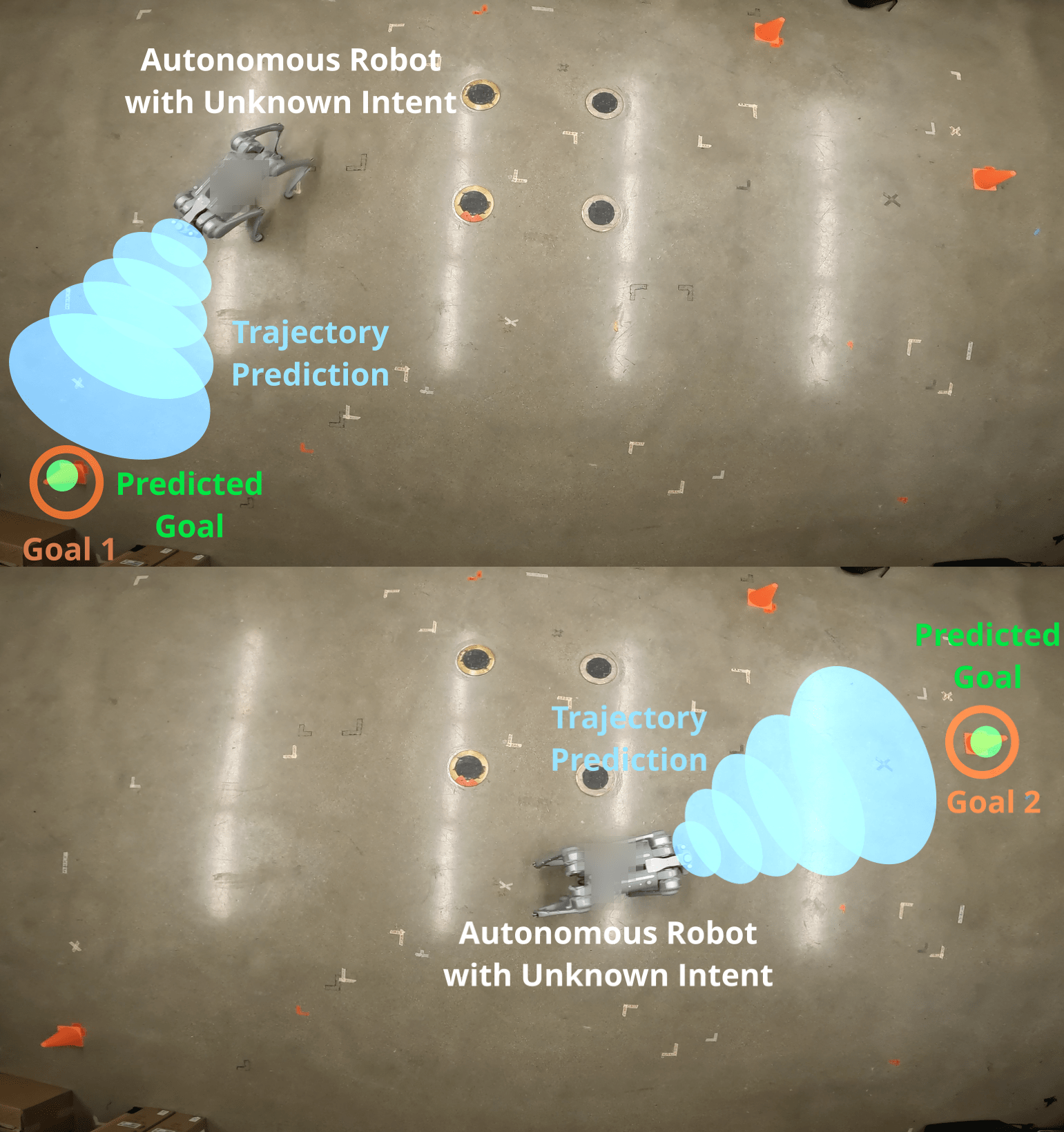}
\caption{Proposed algorithm predicts an autonomous robot's unknown intent (goal) and future trajectory.}
\label{fig:abstract}
\end{figure}
Intention inference and trajectory prediction have been critical research areas for a wide range of autonomous applications. The field has seen significant development within Human-Robot Interaction (HRI). Wang et al.~\cite{wang2013probabilistic} introduced probabilistic movement modeling to infer human intent from observed actions. Within this domain, researchers have explored online adaptation using movement primitives and multimodal Bayesian fusion to reduce uncertainty during collaborative tasks~\cite{trick2019multimodal,koert2019learning}. These HRI-focused methods often rely on the assumption of a cooperative target or the availability of pre-defined motion templates, which may not hold in broader surveillance and security contexts.

Expanding beyond collaborative settings, intention inference is increasingly vital for monitoring non-cooperative agents in complex environments. For instance, \cite{liang2021detection} addressed the detection of malicious intent in drone surveillance, particularly for protecting restricted areas such as airports from non-cooperative intruders. Similarly, in the aerospace domain, \cite{chen2025orbital} investigated intention recognition for non-cooperative targets in orbital environments, where agents may exhibit unpredictable maneuvers to achieve unknown objectives. These surveillance scenarios encounter a fundamental challenge: targets in such settings often behave in a non-cooperative or even adversarial manner, lacking observable signals of their intent. Furthermore, the specific motion characteristics of how a target adheres to a path deterministically are typically unknown or subject to change.

Learning-based approaches have recently shown promise for intention inference and trajectory prediction, employing deep (sequence) models that are trained end-to-end on large, labeled data sets (e.g., LSTM~\cite{alahi2016social, xu2019predicting}, GAN~\cite{gupta2018social}, attention-based network~\cite{Li2024Interactive, xu2019predicting, Liao2025ASafety,chen2025orbital}). Although these methods can capture subtle behaviors, they suffer from two practical limitations: (i) substantial offline training data are required, which are usually unavailable or impractical to collect for novel targets, and (ii) the learned policy is tightly coupled to the environment layout, so the model must be retrained whenever the scene geometry or agent class changes ~\cite{rudenko2020human}. In response to the first challenge, informed machine learning integrates prior knowledge, such as physical laws~\cite{karniadakis2021physics} or structural constraints~\cite{liang2025online}, into the learning process. This strategy can accelerate convergence, relax data requirements, and improve robustness to data noise. To deal with the second challenge, recent studies have explored meta-learning, domain adaptation, and test-time adaptation strategies that aim to improve cross-scene transferability and rapid adaptation to new environments~\cite{ivanovic2022expanding, li2024metatra}. These approaches improve robustness to distribution shifts by learning transferable representations or adaptation mechanisms from multiple training domains. Methods from both directions remain limited by patterns encountered during training or by the assumptions encoded in their priors. Consequently, their ability to generalize to previously unseen behaviors or intentions at runtime is often restricted.

Bayesian approaches\cite{yepes2007new, agamennoni2012estimation, kooij2019context} offer an alternative to learning-based methods by maintaining an explicit belief over candidate goals and updating this belief online. These methods typically employ geometric likelihoods derived from roadmaps or cost-to-go functions \cite{gan2020modeling, zanchettin2017probabilistic}. A particularly well-studied Bayesian framework is the bridging distribution method \cite{ahmad2015destination}, which estimates future trajectories and arrival times by constructing a stochastic process conditioned on each potential destination. By introducing a pseudo-observation at the goal, this method formulates a likelihood function that respects the Markovian structure of the observed trajectory, enabling destination inference through a probabilistic bridge model \cite{ahmad2019bayesian}. Instead of dealing with a single dynamic motion model\cite{petrich2013map, coscia2018long}, using Bayesian frameworks to predict future trajectories is often associated with inference of multiple potential goals. To solve this issue, the Gaussian mixture model (GMM) is used to combine the probability of different goal inferences or trajectories\cite{Kanazawa2019Adaptive, Liao2025ASafety}. 

An alternative line of work frames the problem from a path planning perspective \cite{Rosmann2017Online,karasev2016intent,vasquez2016novel}. For instance, \cite{best2015bayesian}, \cite{yoon2021predictive} assume that the target attempts to follow the shortest path toward its goal, using a Boltzmann-like stochastic kinematic model for state transitions. This assumption allows for a tractable likelihood function over goal hypotheses, conditioned on the observed trajectory. Future state prediction is performed through Monte Carlo sampling, avoiding the need for complex analytical derivations that grow exponentially with the planning horizon.
Despite their strengths, many Bayesian filters \cite{best2015bayesian,ahmad2019bayesian,liang2019destination,gan2020modeling} assume that the target sticks to a single, fixed goal, which often fails in practice when agents re-plan in response to changing objectives or perform deceiving behavior. To handle the potential change of the goal, a forgetting mechanism may be introduced in processing the past data, but it needs to tune a hyperparameter that governs the forgetting speed~\cite{Kanazawa2019Adaptive}. Furthermore, the requirement of prior knowledge for the target’s dynamics is often unavailable or inaccurate in non-cooperative or adversarial settings~\cite{schwarting2019social}.
To address these limitations, we propose a dynamic Bayesian algorithm in which the target’s goal is modeled as a Markovian latent state that can switch arbitrarily over time. Simultaneously, we estimate an intention parameter, which governs the determinism of the target's motion. This parameter is commonly employed to model uncertain or unknown motion dynamics in probabilistic intention inference \cite{hu2021novel, yang2020assisted}, yet it is typically treated as a fixed, predefined quantity in prior work. In contrast, our approach treats the intention parameter as an unknown and adapts its estimate online.

The proposed intention inference and trajectory prediction algorithm in this paper consists of two main components. The first is an adaptive intention inference module that jointly estimates both the current goal of the target and the corresponding intention parameter. These two estimations evolve jointly, influencing each other at each time step to support more accurate and responsive inference. The second component is a sampling-based trajectory prediction module. Rather than deriving an analytical expression for the future state distribution, which becomes analytically intractable due to the combinatorial explosion of possible goal transitions, we simulate multiple trajectory rollouts. These samples are weighted by the current posterior belief over candidate goals, following a strategy similar to \cite{best2015bayesian}. While \cite{best2015bayesian} assumes a fixed goal and intention parameter, our proposed method can produce more accurate predictions by inferring intentions under goal switching and updating the estimate of the intention parameter.

The key contributions of this paper are as follows:
1) We formulate trajectory prediction via intention inference as a dynamic Bayesian inference problem, incorporating a Markovian model to capture potential changes in the target’s intention over time.
2) We propose an adaptive estimation strategy for the unknown intention parameter $\alpha$, allowing the algorithm to adjust its prediction model based on observed behavior dynamically.
3) We develop an algorithm that integrates intention inference and trajectory prediction in real time, operating at approximately \SI{547}{Hz} on the experimental platform. The effectiveness and robustness of the proposed method are validated through extensive simulations and hardware experiments on both quadrotor and quadrupedal robots, as illustrated in Fig.~\ref{fig:abstract}. Comparative results demonstrate that our approach consistently outperforms the baseline and alternative Bayesian methods in terms of accuracy, robustness, and adaptability to dynamic intention changes.

The structure of the paper is as follows. Section \ref{sec: problem_formulation} formulates the problem of interest. Section \ref{sec: algorithm} presents the proposed online intention inference and prediction algorithm. Section \ref{sec: numerical_experiments} evaluates the algorithm's effectiveness through numerical experiments under various challenging scenarios, including the change of the target's intention and unknown dynamics. Section \ref{sec: experiment} extends the evaluation to real-world hardware platforms, including a quadrotor drone and quadrupedal robot, demonstrating the algorithm’s practical applicability.
Section \ref{sec: conclusion} concludes the work and presents future directions.

\section{Problem Formulation}\label{sec: problem_formulation}
Consider a moving target with state denoted by an $n$-dimensional random vector $X_k$ at time $k=0,1,2,...$. In this problem, we address dual uncertainties: the target's intended destination and its specific motion characteristics. Suppose at each time $k$, the target is associated with a \emph{goal state}, which the target's state intends to reach in the future, denoted by an $n$-dimensional random vector $\boldsymbol{\theta}_k$. 

\subsection{Probabilistic Kinematic Model}
We assume the target's state is Markovian and evolves through a transition probability conditioned on the goal state $\boldsymbol{\theta}$. Let $\delta(\cdot, \cdot) $ denote the lowest cost between two points. The transition probability is assumed to adopt a Boltzmann distribution:
{\small
\begin{subequations} \label{eq: target_kinematics}
\begin{align}
\Pr( X_{k+1} &= \boldsymbol{x}_{k+1} \mid X_k = \boldsymbol{x}_k, \boldsymbol{\theta}_k = \boldsymbol{\theta}_k^*) \\
&= \frac{1}{K} \exp[ -\alpha  \bar{\delta}( \boldsymbol{x}_{k+1}, \boldsymbol{x}_{k}, \boldsymbol{\theta}_k^*) ], \nonumber \\
K &= \sum_{\boldsymbol{x}_{k+1} \in \mathcal{X}_k^+} \exp[ -\alpha  \bar{\delta}( \boldsymbol{x}_{k+1}, \boldsymbol{x}_{k}, \boldsymbol{\theta}_k^*) ], \\
\bar{\delta}( \boldsymbol{x}_{k+1}, \boldsymbol{x}_{k}, \boldsymbol{\theta}_k^*) &= \delta(\boldsymbol{x}_k, \boldsymbol{x}_{k+1}) +\delta(\boldsymbol{x}_{k+1}, \boldsymbol{\theta}_k^*) - \delta(\boldsymbol{x}_k, \boldsymbol{\theta}_k^*).
\end{align}
\end{subequations}
}%
where $ \boldsymbol{x}_k \in \mathcal{X} \subset \mathbb{R}^n$ is the value of $X_k$ at time $k$; $\mathcal{X} $ is a discrete set of feasible states; $ \boldsymbol{\theta}_k^*$ denotes the value taken by $\boldsymbol{\theta}_k$ at time $k$.
$K$ is a normalization constant, and $\overline{\delta}$ represents the cost associated with the transition toward the goal. The set $ \mathcal{X}_k^+ \subset \mathcal{X} $ denotes all feasible states reachable from $ \boldsymbol{x}_k $ at time $k$. In the case where $\boldsymbol{x}_{k}$ denotes the position of the target, $\delta$ is the shortest path between the two points. While in more complex cases, $\delta$ can be the cost function of traveling from one point to another; Central to this model is the parameter $\alpha$ (often termed inverse temperature). This parameter governs the determinism of the target's motion. A large $\alpha$ represents a target following an efficient, nearly deterministic shortest-path policy. While a small $\alpha$ represents a target with high stochasticity or ``noisy'' motion, likely deviating from the optimal path. The adoption of this Boltzmann distribution is a common practice in trajectory prediction, as it allows for the probabilistic modeling of kinematics without the need to explicitly specify the target’s underlying dynamics~\cite{hu2021novel, yang2020assisted}.

\subsection{Latent Intentions and Unknown Dynamics}
To handle realistic, non-cooperative scenarios, we introduce two key assumptions.

\begin{assumption}[Varying Goals] \label{assume:1}
The target's intention is not static. The sequence of goal states $\{\boldsymbol{\theta}_{k}\}_{k=0}^{\infty}$ is assumed to be a time-homogeneous Markov process.
\end{assumption}

Assumption~\ref{assume:1} allows the target to switch abruptly between a set of candidate goals $\mathcal{O} \triangleq \{\boldsymbol{o}_{i}, i=1,2,...,N\}$ according to a transition matrix $H$, where $H_{ij}=Pr(\boldsymbol{\theta}_{k+1}=\boldsymbol{o}_{j}|\theta_{k}=\boldsymbol{o}_{i})$.

\begin{assumption}[Unknown Kinematics] \label{assume:2}
The kinematic parameter $\alpha$ is unknown and treated as a persistent characteristic of the target. Specifically, $\alpha$ is assumed to remain constant over the observed trajectory, and its value is not provided a priori.
\end{assumption}

\begin{remark}
Although real targets may exhibit time-varying kinematic characteristics, we interpret $\alpha$ as a persistent attribute of the target's intention over a finite time horizon. This assumption balances modeling fidelity and tractability, enabling reliable online inference of $\alpha$ jointly with the goal $\theta_k$ without requiring an explicit dynamics model for $\alpha$. Importantly, this formulation remains effective in scenarios with time-varying intentions by allowing $\alpha$ to be re-inferred across successive trajectory segments. Such cases are explicitly evaluated in both the simulation and hardware experiments presented in this paper. Extensions to explicitly time-varying parameters $\alpha_k$ can be incorporated by introducing a suitable dynamics model, which we leave for future work.
\end{remark}

\subsection{Problem of interest}
Given a sequence of observed states $X_{0:k} = \{\boldsymbol{x}_0, \boldsymbol{x}_1,$ $ \dots, \boldsymbol{x}_k\}$, the problem of interest is to recursively estimate the joint posterior of the goal and the intention parameter to predict the future state distribution $Pr(X_{k+1:k+T}|X_{0:k})$ over a horizon $T$.\footnote{For notational simplicity, the conditioning terms $\boldsymbol{x}_{0:k}$ and $\boldsymbol{o}_i$ are omitted in the following sections unless needed for clarity.}

\begin{remark}
Modeling the goal state as a time-varying random process is essential in realistic scenarios where the target may switch to a new destination abruptly, which can happen within a short time that may not be immediately detectable from the observed trajectory. This is particularly important in non-cooperative or adversarial settings. Prior Bayesian intention inference frameworks \cite{best2015bayesian, ahmad2015destination, liang2019destination} often assume a static goal, which limits their effectiveness in such dynamic environments. In contrast, introducing a Markovian goal process allows the algorithm to track and predict the evolving intention adaptively, where the goal is not fixed or known in advance
\end{remark}

\section{Algorithm}\label{sec: algorithm}
\begin{figure}
\centering
\includegraphics[width=0.4\textwidth]{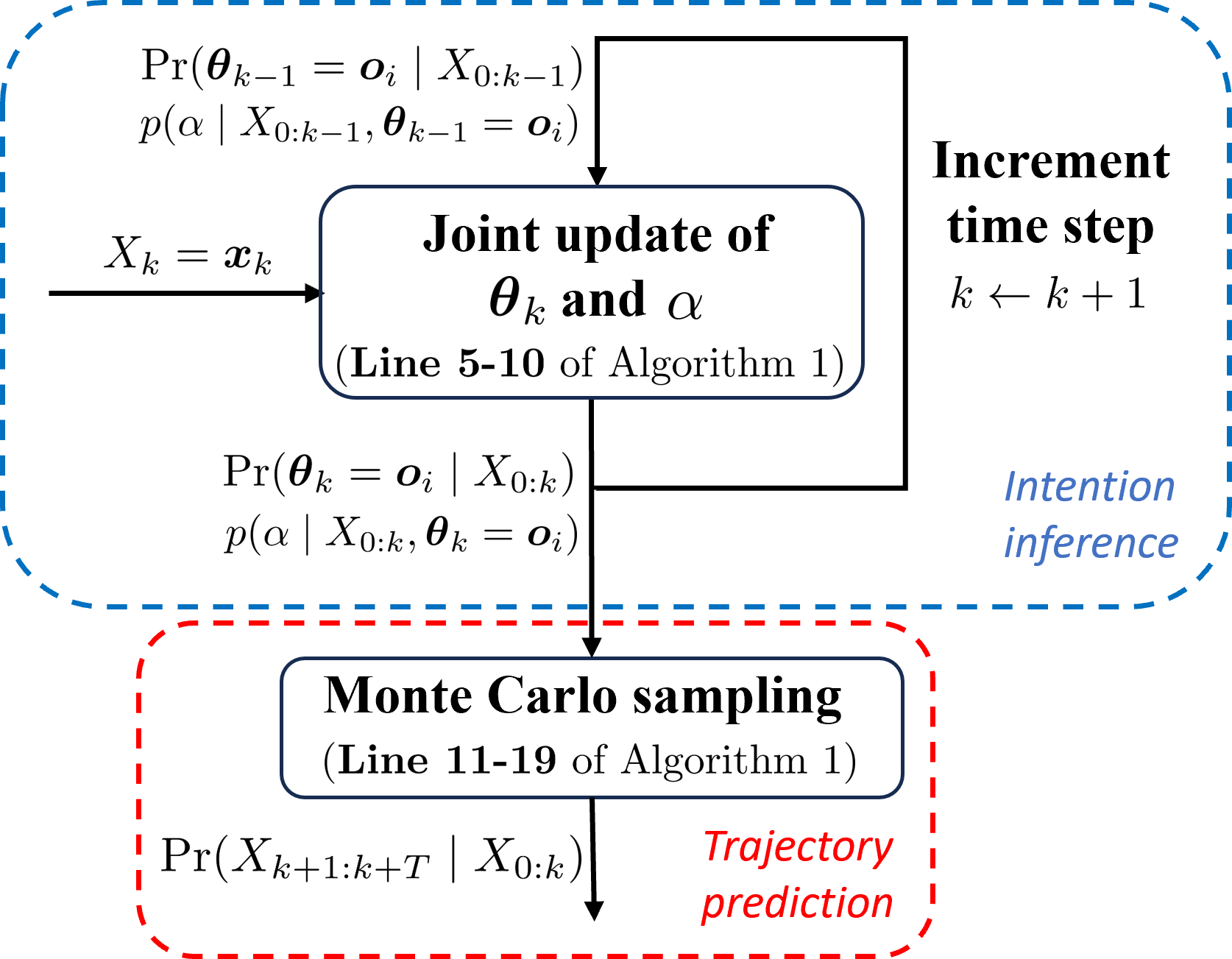}
\caption{Algorithm flow of proposed method}
\label{fig: algorithm_flow}
\end{figure}

The primary challenge in predicting the trajectory of a non-cooperative agent lies in the coupling of intention and dynamics. We proposed a recursive Bayesian algorithm that treats the target's goal and motion characteristics as mutually dependent variables. The overall algorithm flow, summarized in Fig.~\ref{fig: algorithm_flow}, operates as a two-stage process: an intention inference loop followed by a sampling-based trajectory prediction.
\subsection{Inference with Varying Goals and Unknown Kinematics}\label{sec: main_results_inference}
This part performs a joint update of the discrete goal state $\boldsymbol{\theta}_k$ and the fixed intention parameter $\alpha$. We break the recursive update into three logical phases.
\subsubsection{Belief Prediction (Time Evolution)}
Before a new observation $X_k$ is received, we must account for the possibility that the target shifted its goal since the previous step. Using the Markovian transition matrix $H$, the predicted belief for each goal $\boldsymbol{o}_i$ is updated by marginalizing over all possible previous goal states:
\begin{equation}
\label{eq: goal_evolution}
\resizebox{0.4\textwidth}{!}{ 
    $\displaystyle 
    \Pr(\boldsymbol{\theta}_{k}=\boldsymbol{o}_{i}\mid X_{0:k-1}) = \sum_{j=1}^{N} \boldsymbol{H}_{ji} \Pr(\boldsymbol{\theta}_{k-1}=\boldsymbol{o}_{j}\mid X_{0:k-1})
    $
}
\end{equation}
Similarly, the distribution of the intention parameter $\alpha$ is propagated. Because $\alpha$ characterizes the target's adherence to a specific goal, its prior distribution at time $k$ is weighted by the probability of the goal transitions:
\begin{equation}\label{eq: effect_evolution_alpha}
{\small
    \begin{aligned}
    &p(\alpha\mid \boldsymbol{\theta}_{k}=\boldsymbol{o}_{i}, X_{0:k-1})\\
   &=\sum_{j}^{N} p(\alpha\mid\boldsymbol{\theta}_{k}=\boldsymbol{o}_{i}, \boldsymbol{\theta}_{k-1} = \boldsymbol{o}_{j}, X_{0:k-1}),\\
   &\Pr(\boldsymbol{\theta}_{k-1} = \boldsymbol{o}_{j}\mid \boldsymbol{\theta}_{k}=\boldsymbol{o}_{i}, X_{0:k-1})\\
   &=\sum_{j}^{N} p(\alpha\mid \boldsymbol{\theta}_{k-1} = \boldsymbol{o}_{j}, X_{0:k-1}) \frac{\boldsymbol{H}_{ji}\Pr(\theta_{k-1} = \boldsymbol{o}_{j}\mid  X_{0:k-1})}{\Pr(\theta_{k}=\boldsymbol{o}_{i}\mid  X_{0:k-1})}.
\end{aligned}
}
\end{equation}
\subsubsection{Marginal Likelihood Computation}
When the observation $X_k = \boldsymbol{x}_k$ arrives, we evaluate its alignment with each goal $\boldsymbol{o}_i$. Since $\alpha$ is unknown, we compute a marginal likelihood by integrating over our current distribution of $\alpha$:
\begin{equation}\label{eq: likelyhood_states}
{\small
    \begin{aligned}
    \Pr( X_{k}\mid  X_{0:k-1},\boldsymbol{\theta}_{k})
    =&\int \Pr( X_{k}\mid X_{k-1},\boldsymbol{\theta}_{k},\alpha)\\&p(\alpha\mid X_{0:k-1},\boldsymbol{\theta}_{k})\mathrm{d}\alpha.
\end{aligned}
}
\end{equation}
This step allows the algorithm to remain robust against unknown motion dynamics by weighting the likelihood based on the target's inferred motion characteristics.
\subsubsection{Recursive Posterior Update}
Finally, we apply Bayes' rule to obtain the updated joint belief. The goal posterior is updated as follows,\footnote{Using the Markovian property of states, $\Pr( X_{k}\mid  X_{k-1},\boldsymbol{\theta}_{k}) = \Pr( X_{k}\mid  X_{0:k-1},\boldsymbol{\theta}_{k})$.}
\begin{equation}\label{eq: update_theta}
{\small
\Pr(\boldsymbol{\theta} \mid  X_{0:k}) \propto \Pr(\boldsymbol{\theta} \mid  X_{0:k-1})\Pr( X_k \mid  X_{k-1}, \boldsymbol{\theta}).
}
\end{equation}
Simultaneously, the distribution for $\alpha$ is refined for each goal candidate:
\begin{equation}\label{eq: update_alpha}
{\small
    \begin{aligned}
    p(\alpha\mid X_{0:k},\boldsymbol{\theta}_{k})
    =p(\alpha\mid X_{0:k-1},\boldsymbol{\theta}_{k})
    \frac{\Pr( X_{k}\mid X_{0:k-1},\boldsymbol{\theta}_{k}, \alpha)}{\Pr( X_{k}\mid X_{0:k-1},\boldsymbol{\theta}_{k})}.
\end{aligned}
}
\end{equation}
\subsection{Prediction with Unknown Motion Characteristics}\label{sec: main_results_prediction}
To predict the future trajectory one step further $\Pr(X_{k+1}\mid X_{k})$, we marginalize the probability road map \eqref{eq: target_kinematics} with the current inference probability $\Pr(\boldsymbol{\theta}_k\mid X_{0:k})$,
\begin{equation}\label{eq: update_states}
{\small
\begin{aligned}
&\Pr( X_{k+1} \mid  X_{0:k})=\sum_{\boldsymbol{o}_{i} \in \mathcal{O}}\Pr(X_{k+1} \mid  X_{k}, \boldsymbol{\theta}_{k})\Pr(\boldsymbol{\theta}_{k}\mid  X_{0:k}).
\end{aligned}
}
\end{equation}
$\Pr(X_{k+1} \mid  X_{k}, \boldsymbol{\theta}_{k})$ in \eqref{eq: update_states} can be computed in the same way as in \eqref{eq: likelyhood_states}. For future trajectory prediction, this model is iteratively computed for all possible unobserved trajectories in future time steps,
\begin{equation}
{\small
\begin{aligned}\label{eq: trajcetory_prediction_analytical}
\Pr( X_{k+q} \mid  X_{0:k})
&=\sum\Pr( X_{k+q} \mid  X_{k+q-1})\\&\Pr( X_{k+q-1} \mid  X_{0:k}),
\end{aligned}
}
\end{equation}
where the summation is computed over $X_{k+q-1} \in \mathcal{X}_{k+q}^-$ and $\mathcal{X}_{k+q+1}^{-}$ denote the set of states that can reach $\mathcal{X}_{k+q+1}$ in one time step. Rather than deriving an analytical expression for the future state distribution $\Pr(X_{k+1:k+T} \mid X_{0:k})$, which becomes intractable due to the combinatorial explosion of goal transitions and time steps, we employ a Monte Carlo sampling strategy.
\subsubsection{Conditional Sampling}
At each time step $k$, we compute the expectation or mode $\hat{\alpha}_{k,i}$ for each goal candidate $o_i$. The total number of simulation samples $M$ is divided among goals such that $M_i \propto Pr(\theta_k = o_i | X_{0:k})$. By simulating $M$ trajectories starting from the current state $x_k$ using the transition model in \eqref{eq: target_kinematics}, we collect a set of predicted future states for each step $q$ in the horizon $T$.
\subsubsection{Spatial Discretization and Bin Strategy}
To represent the continuous future state distribution $Pr(X_{k+q} | X_{0:k})$ in a computationally efficient and interpretable manner, we employ a spatial binning strategy \cite{best2015bayesian}:
The environment is discretized into a grid of state bins. For each time step $q$ within the prediction horizon, we record the frequency with which the $M$ simulated trajectories visit each specific bin. This raw count is normalized by the total number of samples $M$ to derive a spatial probability distribution. The resulting distribution allows us to identify the most likely future position (the bin with the maximum probability) while the spread of the occupied bins provides a direct measure of prediction uncertainty.
\subsubsection{The Role of $\alpha$ in Prediction Stochasticity}
It is important to clarify that while $\hat{\alpha}_{k,i}$ is treated as a fixed estimate for the duration of the rollout, the randomness in the predicted trajectories arises from the stochastic transition probability defined in \eqref{eq: target_kinematics}. The parameter $\alpha$ serves to scale this randomness. A higher inferred $\alpha$ produces samples tightly clustered in bins along the shortest path (high determinism), whereas a lower $\alpha$ results in a broader distribution of occupied bins, reflecting higher motion stochasticity.
\subsection{Proposed Algorithm Summary}
The complete logic for the joint intention inference and trajectory prediction is summarized in Algo. \ref{alg: adaptive_alg}. This algorithm is designed for high-frequency applications, where the inference loop and the prediction module operate together to provide continuous probabilistic inference and trajectory prediction.
\subsubsection{Initialization and Setup}
At $k=0$, the algorithm initializes the goal belief $\Pr(\theta_0|X_0)$ and the distribution over $\alpha$. Without prior knowledge, these are typically set to uniform and Gamma distributions, respectively. The cost function $\delta(\cdot, \cdot)$, representing the shortest-path distances between all states, is pre-computed using heuristic methods like A-star or RRT\cite{devaurs2015optimal} to ensure computational efficiency at runtime.
\subsubsection{Inference and prediction}
The algorithm follows the recursive structure outlined in the previous subsections: 1) Inference (Lines 5–10): The algorithm performs the joint update of $\theta_k$ and $\alpha$ using the Marginal Likelihood to account for unknown motion characteristics. 2) Prediction (Lines 11–19): Future trajectories are generated via Monte Carlo sampling. Crucially, the samples are weighted by the latest goal posterior, ensuring that the predicted distributions focus on the most probable destinations. The future state distribution $Pr(X_{k+q} \vert X_{0:k})$ is represented by mapping the simulated trajectories onto a discretized grid of state bins. By counting the frequency with which samples visit each bin at each step $q$ of the horizon $T$, it constructs a normalized spatial probability distribution.

\begin{remark}\label{remark: parallel_compute}
In Algo.~\ref{alg: adaptive_alg}, the \textbf{\textup{parfor}} loops starting at lines 5, 12, 15, and 18 are parallelizable. This parallelism makes the algorithm well-suited for real-time applications. Although parallel implementation is not used in our experiments, the algorithm is still capable of operating in real time. 
\end{remark}

\begin{remark}\label{remark: pred_choice}
In relatively simple environments with an approximately unimodal distribution in future states, the predicted trajectory can be represented by the mean of the Monte Carlo samples, as commonly done for deterministic metrics such as Average Displacement Error (ADE) and Final Displacement Error (FDE). In complex environments with obstacles or multiple feasible paths, however, the distribution may become multi-modal, and the sample mean may be physically infeasible or fail to represent the prediction uncertainty. In such cases, spatial binning captures the multi-modal structure by assigning probability mass to distinct regions associated with different feasible paths.
\end{remark}

\begin{algorithm2e}
\caption{Adaptive Joint Inference and Probabilistic Trajectory Prediction}\label{alg: adaptive_alg}
\DontPrintSemicolon
\KwIn{$k = 1$, prediction horizon $T$, sample count $M$, grid resolution, observation $X_0 \in \mathcal{X}$}
Initialize $p(\alpha \mid X_{0}, \boldsymbol{\theta}_{0})$, $\Pr(\boldsymbol{\theta}_{0} \mid X_{0})$, and pre-compute cost matrix $\delta(\cdot,\cdot)$\;

\While{\textbf{True}}{
    Observe $X_{k}$\;
    
    \uIf{$X_{k} \neq X_{k-1}$}{
        \tcp{Intention Inference}
        $\textbf{par}$\For {$i=1,2,\dots,N$}{
            Predict goal shift: $\Pr(\boldsymbol{\theta}_{k} \mid X_{0:k-1}) \gets$ \eqref{eq: goal_evolution}\;
            Propagate $\alpha$: $p(\alpha \mid X_{0:k-1}, \boldsymbol{\theta}_{k}) \gets$ \eqref{eq: effect_evolution_alpha}\;
            Marginal Likelihood: $\Pr(X_{k} \mid X_{0:k-1}, \boldsymbol{\theta}_{k}) \gets$ \eqref{eq: likelyhood_states}\;
            Update $\alpha$ posterior: $p(\alpha \mid X_{0:k}, \boldsymbol{\theta}_{k}) \gets$ \eqref{eq: update_alpha}\;
            Update goal posterior: $\Pr(\boldsymbol{\theta}_{k} \mid X_{0:k}) \gets$ \eqref{eq: update_theta}\;
        }
        \tcp{Trajectory Prediction}
        $\mathcal{X}_{k+1:k+T} = \emptyset$\;
        {$\textbf{par}$\For {$i=1,2,\dots,N$} {
            Allocate $M_{i} \gets \Pr(\boldsymbol{\theta}_{k} \mid X_{0:k}) \times M$\;
            Extract expectation/mode: $\hat{\alpha}_{k,i} \gets \mathbb{E}[\alpha \mid X_{0:k}, \boldsymbol{\theta}_{k}]$\;
            Simulate rollout: $x_{k+1:k+T} \gets \text{Sampling}(X_{k}, \boldsymbol{o}_{i}, \hat{\alpha}_{k,i})$\;
            $\mathcal{X}_{k+1:k+T} \gets \mathcal{X}_{k+1:k+T} \cup \{x_{k+1:k+T}\}$\;
        }
        }
        $\textbf{par}$\For {$j=1,2\dots,T$}{
            $\Pr(X_{k+j} \mid X_{0:k})\gets\frac{\text{CountBin}(\mathcal{X}_{k+j})}{M}$\; 
        }
    }
    $k \gets k + 1$\;
}
\end{algorithm2e}

\section{Numerical Experiments}
\label{sec: numerical_experiments}

In this section, we evaluate the performance and robustness of the proposed algorithm through a series of comparative experiments and an ablation study. Our primary objective is to verify how joint adaptive estimation of goal intention and motion kinematics improves inference and prediction accuracy under uncertainty. To isolate the contributions of each adaptive component, we compare four distinct methods:

\begin{itemize}
    \item \textbf{(B) Baseline:} A state-of-the-art planning-based algorithm that assumes a fixed goal and static kinematics, requiring no offline training \cite{best2015bayesian}.
    \item \textbf{(A) $\alpha$-Adaptive:} The baseline algorithm incorporating only the proposed adaptive estimation of the intention parameter $\alpha$.
    \item \textbf{(G) Goal-Adaptive:} The baseline algorithm incorporates only the proposed Markovian latent state model for goal estimation.
    \item \textbf{(P) Proposed:} The full algorithm combining both adaptive $\alpha$ and adaptive goal estimation.
\end{itemize}

The ablation study allows us to identify the specific performance enhancements driven by each individual component, particularly regarding rapid adaptation to abrupt goal shifts and improvements in long-term prediction accuracy. We validate these claims through two targeted case studies, a comprehensive 500-trial Monte Carlo simulation to assess statistical robustness, and a scalability test designed to evaluate the algorithm's computational limits as the number of goal candidates and state-space discretization increase. These evaluations demonstrate how the joint estimation of goals and motion characteristics provides favorable results compared to non-adaptive methods.

\subsection{Simulation Setup}\label{subsec: numerical_experiments_setup}

We establish a standardized simulation environment with static obstacles that reflects the challenges of intention inference and trajectory prediction under unknown motion dynamics and complex environments. This setup is designed to test the algorithm's ability to maintain a belief over candidate goals while simultaneously adapting to the target's specific motion characteristics, represented by the intention parameter $\alpha$. Unless specified, the following parameters and configurations are consistently applied across our case studies and statistical analyses to ensure a controlled comparison between the four evaluated algorithms.

\subsubsection{Environment and Algorithm Configuration}

In these numerical experiments, the target moves within a discrete grid of $81 \times 61$ nodes while some of the nodes are blocked by obstacles and are not achievable. The computation of the $\delta$ function is performed using the A-star method to ensure efficient path cost calculation. Candidate goals with $N=45$ are distributed along the boundary of the map, while several are scattered throughout the middle of the workspace to account for more complex navigation scenarios. Each goal candidate is assigned an equal initial probability to avoid prior bias in the inference process. This setup represents a specific example of the general goal candidate assignment strategies and allows for the evaluation of the algorithm's performance in environments where intentions may change in open space.

For algorithms A and P, which adapt $\alpha$ with observations, we initialize the belief over $\alpha$ for each goal with a Gamma distribution, specifically $\Gamma(3,3)$, to reflect the assumption that $\alpha$ is strictly positive. The trajectory prediction uses $M = 500$, $T = 20$, and an overall estimate of the intention parameter $\alpha$. We define this overall estimate $\hat{\alpha}_k = \sum_{i=1}^{N} \hat{\alpha}_{k,i} \Pr(\boldsymbol{\theta}_k=\boldsymbol{o}_i \mid X_{k})$ in algorithms A and P. For methods G and P, which treat the goal states as Markovian states, the transition matrix $\boldsymbol{H}$ of $\boldsymbol{\theta}_k$ is constructed to preserve its Markovian property. The diagonal entries are set to $1 - \frac{1}{0.0025N}$, while the off-diagonal entries are assigned equal values such that each row sums to one. This specific design of $\boldsymbol{H}$ implies that if a target is heading toward its current goal, there is a high probability it will not change its goal in the next step. However, if it does switch goals, it is assumed to have equal probability of transitioning to any of the other goals, as we lack specific prior knowledge. If one believes it is more likely to switch from the $i$-th goal to the $j$-th, the corresponding entry $\boldsymbol{H}_{ij}$ can be set to a larger value.

\begin{remark}
For algorithm A, we assume no evolution of the goal state $\boldsymbol{\theta}$ occurs; intention inference is computed by replacing the time-varying $\boldsymbol{\theta}_k$ with a static $\boldsymbol{\theta}$ in the update laws. Conversely, for algorithm G, the intention parameter $\alpha$ is treated as a fixed, predefined constant, while intention inference follows the proposed Markovian transition logic. The sampling-based trajectory prediction mechanism remains consistent across all four algorithms to ensure a fair comparison.
\end{remark}

\subsubsection{Case 1}

The reference trajectory begins from an initial position and evolves according to the dynamics described in \eqref{eq: target_kinematics}, using a fixed parameter $\alpha^{\ast} = 8$ and a goal state $\boldsymbol{\theta}$. The goal $\boldsymbol{\theta}$ transitions sequentially through the set $\{\boldsymbol{\theta}_{1}, \boldsymbol{\theta}_{2}, \boldsymbol{\theta}_{3}\}$ at specific time steps, ultimately terminating at $\boldsymbol{\theta}_3$. We deliberately set a significant difference between the true intention parameter $\alpha^{\ast}$ and the estimate $\hat{\alpha}=1$ for B and G, specifically $\hat{\alpha} \ll \alpha^{\ast}$, to illustrate the performance degradation caused by mismatched $\alpha$ values.

\subsubsection{Case 2}
The reference trajectory begins from the same initial position as in Case 1 but with $\alpha^{\ast} = 2$. The goal state $\boldsymbol{\theta}$ switches from $\boldsymbol{\theta}_{1}$ to $\boldsymbol{\theta}_{2}$ and $\boldsymbol{\theta}_{3}$ sequentially. For B and G, we fix $\hat{\alpha} = 8$, which is significantly larger than $\alpha^{\ast}$. This case evaluates the ability of the proposed method to maintain prediction accuracy even when the initial motion model is highly overconfident.

\subsubsection{Monte Carlo Simulation}

To evaluate the robustness and statistical significance of the proposed algorithm, we conduct a Monte Carlo experiment consisting of 500 randomly generated target trajectories. In each trajectory, the target updates its intended goal position after a random interval between 30 and 100 time steps, where the new destinations are selected randomly from the set of all possible goal locations. Within each specific segment of the trajectory occurring between intention changes or prior to termination, the true intention parameter $\alpha^{\ast}$ is uniformly sampled from the interval $[0.1, 30]$ to represent a diverse range of motion behaviors.

The four methods evaluated in the previous simulations are applied to perform intention inference and trajectory prediction across the complete set of 500 trajectories. $M=300$ is used in Monte Carlo sampling for the trajectory prediction module. For the baseline methods B and G, a fixed estimate of $\hat{\alpha} = 4$ is assigned, which represents a moderate value between high certainty and high uncertainty.

\begin{figure*} [h]
\subfloat[Inference and prediction at time step 50]
{\label{fig: case_01_step_40} \includegraphics[width=0.32\linewidth]{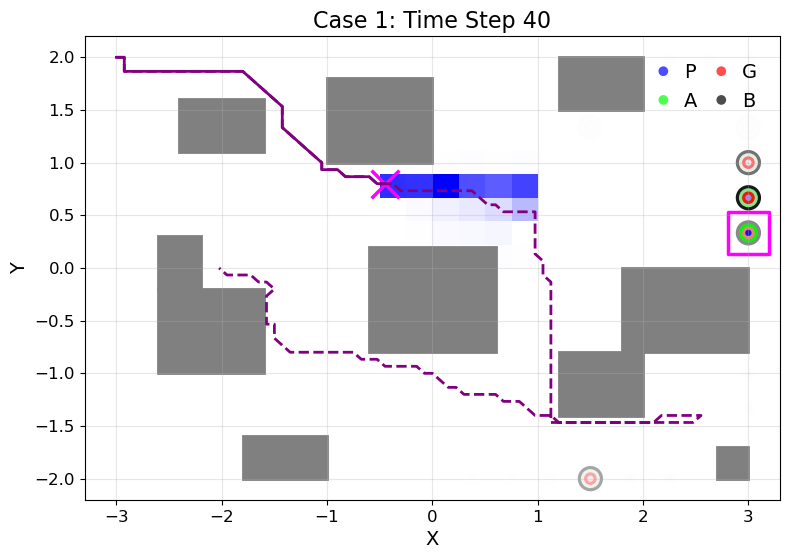}}
\hfill
\subfloat[Inference and prediction at time step 100]
{\label{fig: case_01_step_90} \includegraphics[width=0.32\linewidth]{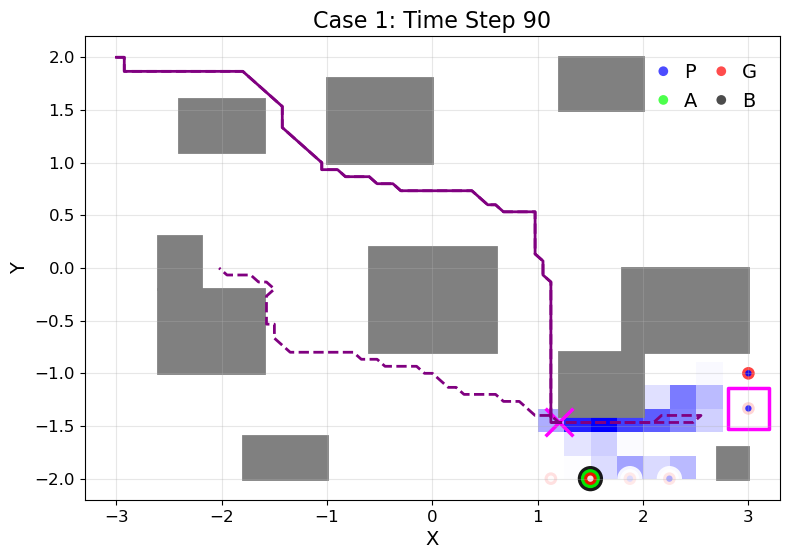}}
\hfill
\subfloat[Inference and prediction at time step 150]
{\label{fig: case_01_step_150} \includegraphics[width=0.32\linewidth]{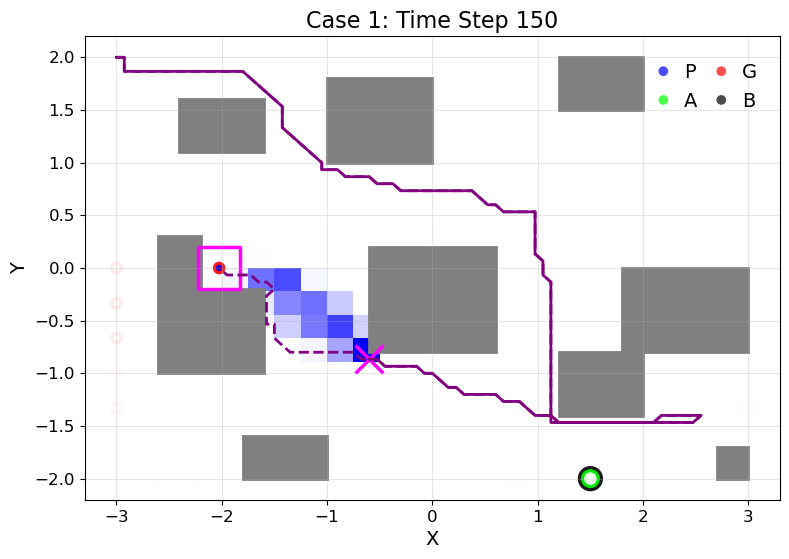}}
\caption{Trajectory visualizations at different time steps for Case 1. Each subplot displays the target's movement alongside the inference and prediction results of the four methods, each in a distinct color. The pink cross indicates the current position, and the pink square marks the true goal. The dashed purple line denotes the reference trajectory, which is unknown to the inference algorithm. Results are color-coded by method. Candidate goals are represented as circles, with darker shading indicating higher inferred probability. Only the predicted future trajectory of the proposed method is illustrated by blue bins, which are normalized within the prediction horizon. Deeper shades show a higher probability of the target visiting this bin in the future.} \label{fig: simulation_case_01}
\end{figure*}

\subsubsection{Scalability Test}

To examine the effects of grid refinement, we evaluate the impact of the total number of potential goals $N$, the prediction horizon $T$, and the number of samples $M$ used in the Monte Carlo trajectory prediction. We conduct a systematic scalability test where each feature is assigned three distinct values to observe performance trends. Each parameter tuple is evaluated using 50 randomly generated trajectories, with half of the goal candidates positioned along the workspace boundary and the remaining half distributed randomly within the middle of the free space. 

It is important to note that direct comparison of the inference probability for the true goal is not statistically meaningful across different grid densities or varying numbers of candidate goals. Consequently, this analysis focuses on trajectory prediction accuracy and computation time to address the requirements of real-world applications. For each specific combination of grid refinement and goal count, the same 50 trajectory sets are utilized to consistently evaluate the performance across varying prediction lengths and sample sizes. This approach reveals the computational limits and ensures that the real-time performance claims remain valid as the environment representation becomes more detailed.

\subsection{Evaluation Metrics}
We employ three quantitative metrics to evaluate both the intention inference performance and the trajectory prediction performance of the proposed algorithm in numerical experiments.

\subsubsection{Inference Metric}
For intention inference, the performance is evaluated using the posterior probability assigned to the true goal: $\Pr\!\left(\boldsymbol{\theta}=\boldsymbol{\theta}_{\mathrm{true}}\mid X_{0:k}\right).$
A larger value indicates that the inference algorithm assigns higher confidence to the correct target intention.
\subsubsection{Trajectory Prediction Metrics}
For each observation time $k$, the algorithm predicts a distribution over spatial bins for each prediction step $t\in\{1,2,\dots,T\}$. Let $p_{i,t}$ denote the predicted probability assigned to bin $i$ at timestep $t$.
\paragraph{Prediction Accuracy (ACC)}
Prediction accuracy evaluates whether the ground-truth target position
falls inside the most probable predicted bin at prediction step $t$.
Let $i_t^{*}=\arg\max_i p_{i,t}$ denote the index of the bin with the highest predicted probability, and
let $i_t^{\mathrm{gt}}$ denote the index of the bin containing the
ground-truth target position $x_{k+t}$. The prediction accuracy at
prediction step $t$ is defined as
\vspace{-3pt}
\begin{equation}
\mathrm{ACC}_t
=
\mathbb{I}\!\left(i_t^{*}=i_t^{\mathrm{gt}}\right),
\label{eq:acc_metric}
\end{equation}
where $\mathbb{I}(\cdot)$ is the indicator function.
\paragraph{Negative Log-Likelihood (NLL)}
While ACC only evaluates the most likely predicted bin, it does not account for the overall predicted probability distribution. Therefore, we additionally evaluate the prediction quality using the negative log-likelihood (NLL):
\vspace{-3pt}
\begin{equation}
\mathrm{NLL}_t=-\log\!\left(p_{i_t^{\mathrm{gt}},t}\right),
\label{eq:nll_metric}
\end{equation}
A lower NLL indicates that the predicted probability distribution assigns higher likelihood to the ground-truth trajectory and therefore reflects better probabilistic prediction performance. In the following evaluations, both ACC and NLL are averaged over the prediction horizon $T$ to summarize the trajectory prediction performance at observation time $k$.

\subsection{Adaptation to Abrupt Intention Shifts}\label{subsec: numerical_experiments_adpt_goal}

\begin{figure*}
\subfloat[Inference probability of true goal in Case 1 (higher is better).]
{\label{fig: case_01_infer_plot} \includegraphics[width=0.32\linewidth]{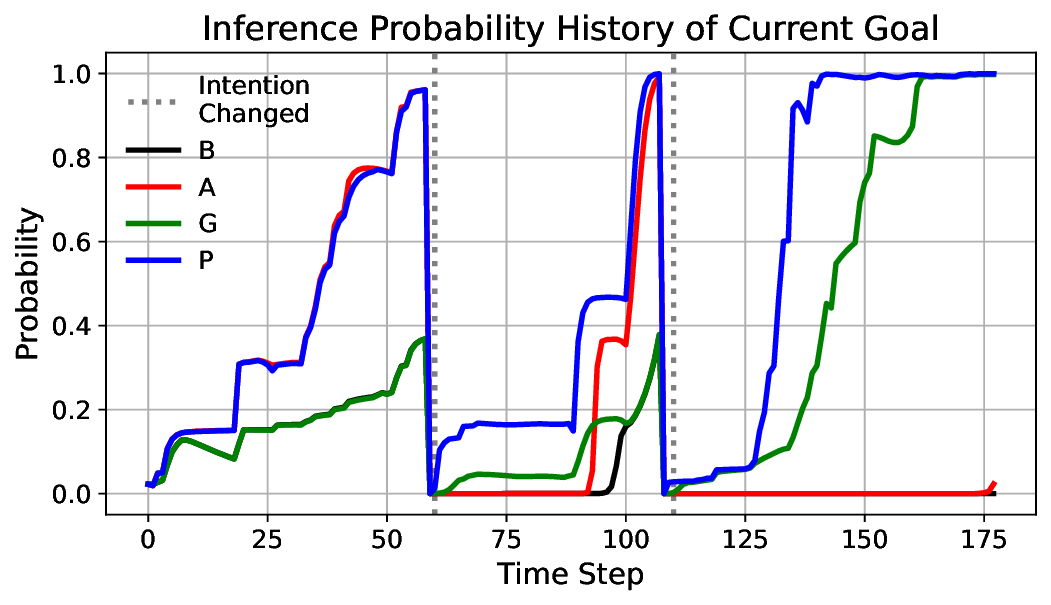}}
\hfill
\subfloat[Prediction ACC in Case 1 (higher is better).]
{\label{fig: case_01_pred_acc_plot} \includegraphics[width=0.32\linewidth]{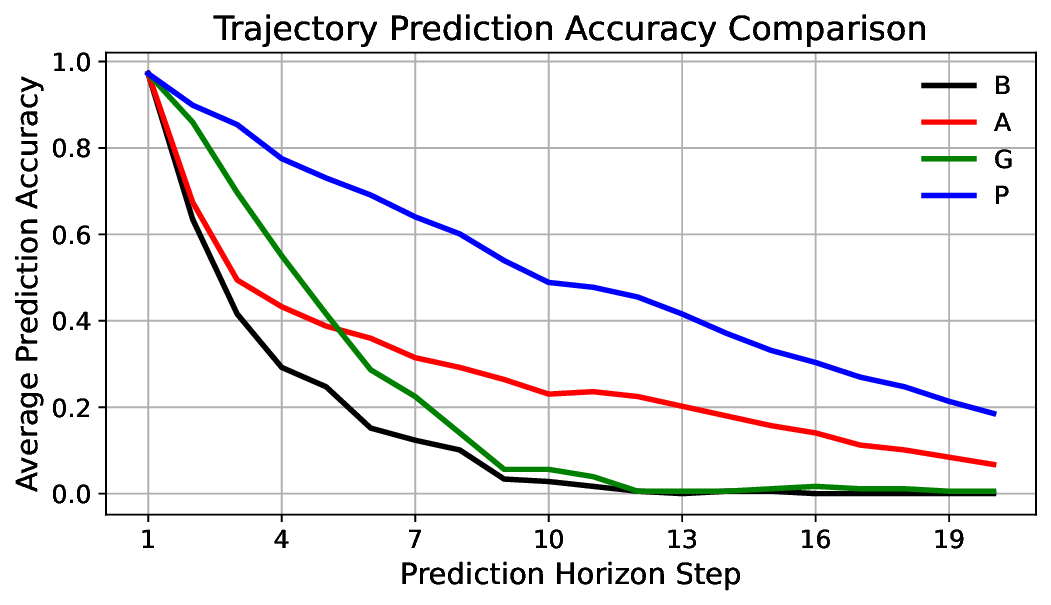}}
\hfill
\subfloat[Prediction NLL in Case 1 (lower is better).]
{\label{fig: case_01_pred_nll_plot} \includegraphics[width=0.32\linewidth]{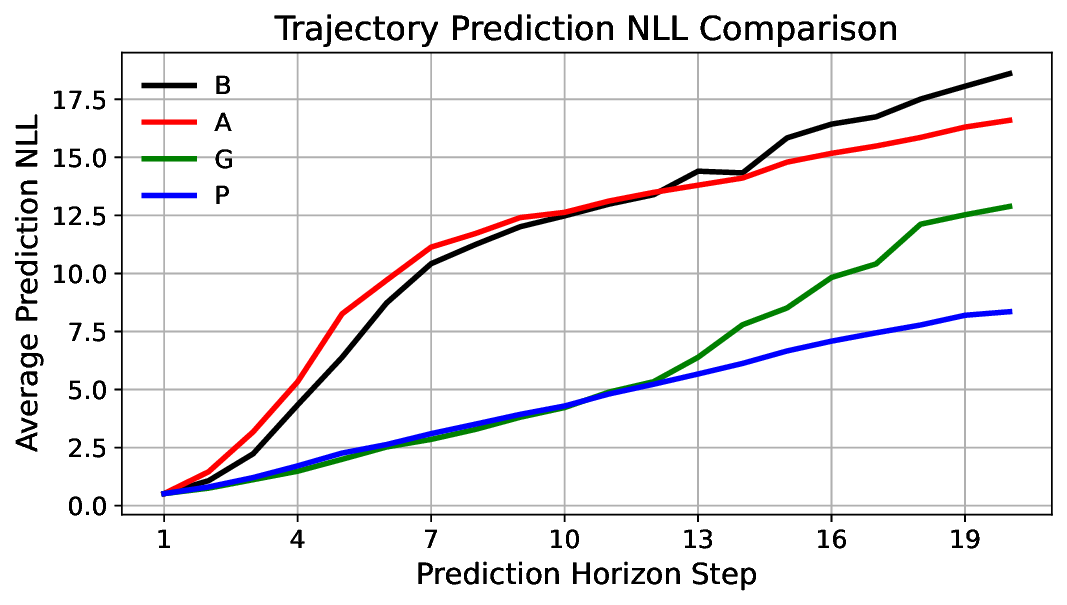}}
\caption{Evaluation metrics in Case 1. The proposed methods outperform other baselines in each metric} \label{fig: simulation_case_01_inf}
\end{figure*}

A primary strength of the proposed algorithm is its responsiveness to abrupt shifts in target intent. As illustrated in Fig.~\ref{fig: simulation_case_01}, the target's reference trajectory sequentially targets three distinct goals. Initially, all four algorithms successfully infer the first goal located on the left of the environment; the inference probability of the true goal rises steadily for each method during this phase, as shown in Fig.~\ref{fig: case_01_infer_plot}.

However, the performance diverges significantly following the first goal transition (occurring between time steps 60 and 90). While the non-adaptive Baseline (B) and $\alpha$-Adaptive (A) methods fail to react to this shift, the Goal-Adaptive (G) and Proposed (P) methods successfully capture the change, rapidly increasing the inference probability for the second goal (see Fig.~\ref{fig: case_01_step_90} and Fig.~\ref{fig: case_01_infer_plot}).

This trend becomes even more pronounced when the target switches to its third goal. Algorithms B and A incorrectly infer the true goal to be in the bottom-right of the environment, which is the opposite direction of the actual motion. In contrast, algorithms G and P, leveraging the adaptive goal module, correctly identify the true goal (see  Fig.~\ref{fig: case_01_step_150}). Modeling the goal state as a Markovian process allows the algorithm to quickly recover from and react to shifts in intent, as evidenced by the sharp increase in inference probability immediately following the dashed transition lines in Fig.~\ref{fig: case_01_infer_plot}.

As illustrated in Fig.~\ref{fig: case_01_pred_acc_plot}, all evaluated algorithms exhibit high accuracy at the onset of the prediction horizon. This is expected, as predictions are initialized from the current observed state, and initial samples are naturally clustered in bins surrounding the starting position. As the prediction horizon increases, accuracy typically declines for all methods due to the accumulation of stochastic uncertainty. However, the proposed algorithm consistently maintains the highest accuracy across the horizon, demonstrating the benefit of joint goal and parameter adaptation in reducing long-term divergence.

This advantage is further reflected in the negative log-likelihood (NLL) results shown in Fig.~\ref{fig: case_01_pred_nll_plot}. The proposed method consistently maintains the lowest NLL with a gradual and approximately linear increase throughout the prediction horizon, indicating that its predicted probability distribution remains robust under abrupt goal changes. In contrast, the non-adaptive methods exhibit substantially larger NLL values, with the degradation becoming more pronounced at later prediction steps. 

\subsection{Robustness to Unknown Motion Characteristics}\label{subsec: numerical_experiments_robust_alpha}

\begin{figure}[htb]
\centering
\subfloat[Error between $\hat{\alpha}$ and $\alpha^{\ast}$ in Case 1 (lower is better).]
{\label{fig: case_01_alpha_plot} \includegraphics[width=0.85\linewidth]{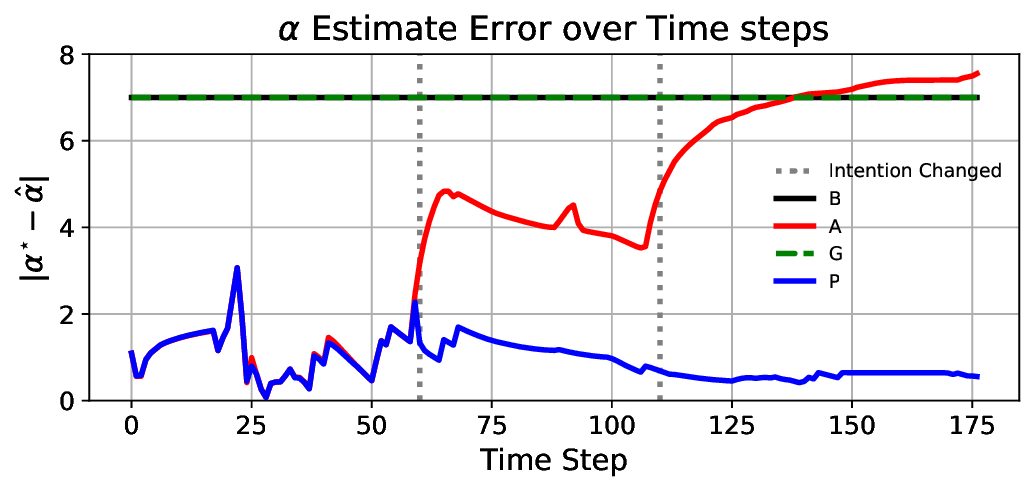}}\\[1ex]
\subfloat[Error between $\hat{\alpha}$ and $\alpha^{\ast}$ in Case 2 (lower is better).]
{\label{fig: case_02_alpha_plot} \includegraphics[width=0.85\linewidth]{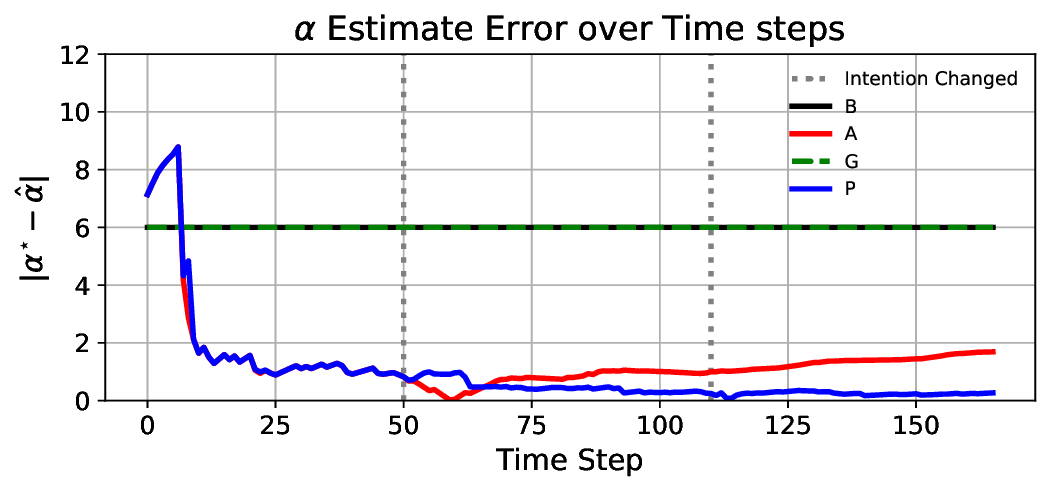}}
\caption{$\alpha$ estimation error in Case 1 and 2} \label{fig: simulation_case_02}
\end{figure}

Another advantage of the proposed algorithm is its robustness to unknown motion characteristics, achieved through the online estimation of the intention parameter $\alpha$ within the probabilistic roadmap defined in \eqref{eq: target_kinematics}. As illustrated in Fig.~\ref{fig: case_01_alpha_plot} and Fig.~\ref{fig: case_02_alpha_plot}, the Proposed (P) algorithm yields significantly more accurate parameter estimates than non-adaptive methods. While algorithms B and G maintain a large, static estimation error due to their fixed $\hat{\alpha}$, the adaptive algorithms (A and P) demonstrate the ability to reduce this error over time. Crucially, algorithm P consistently converges toward the true value $\alpha^{\ast}$ regardless of intention shifts. In contrast, algorithm A, which lacks goal-switching logic, fails to maintain accuracy during goal transitions because it cannot decouple motion stochasticity from intention changes.

\begin{figure*}[htb]
\subfloat[Inference probability of true goal in Case 2 (higher is better).]
{\label{fig: case_02_infer_plot} \includegraphics[width=0.32\linewidth]{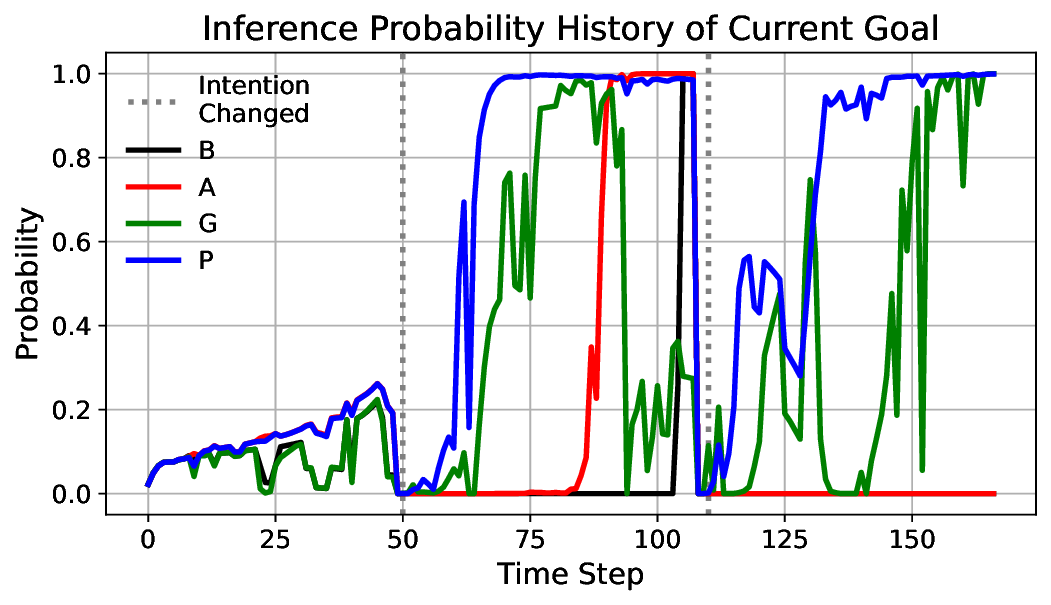}}
\hfill
\subfloat[Prediction ACC in Case 2 (higher is better).]
{\label{fig: case_02_pred_acc_plot} \includegraphics[width=0.32\linewidth]{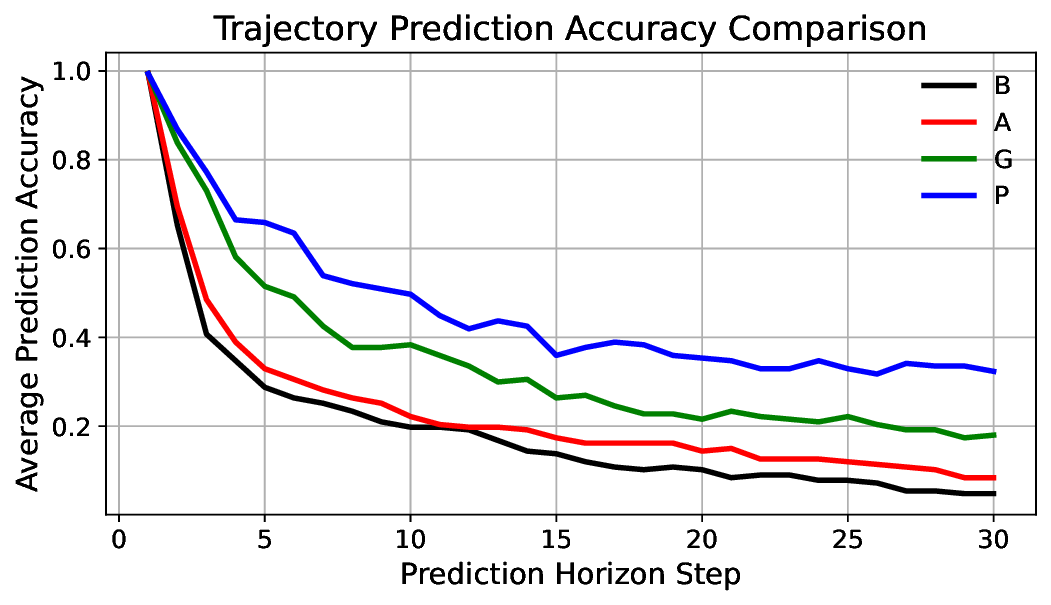}}
\hfill
\subfloat[Prediction NLL in Case 2 (lower is better).]
{\label{fig: case_02_pred_nll_plot} \includegraphics[width=0.32\linewidth]{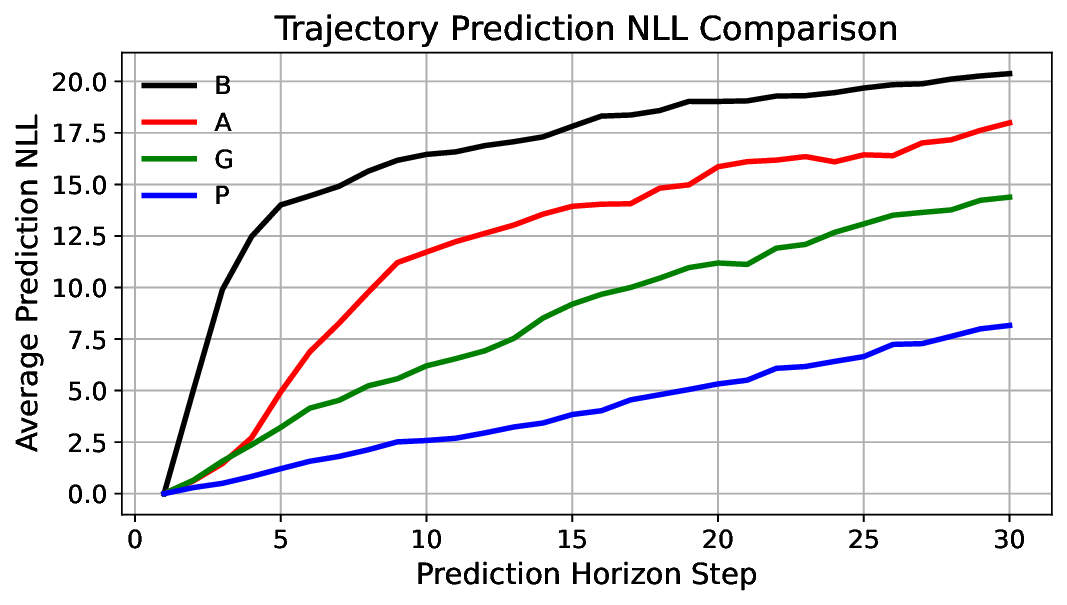}}
\caption{Evaluation metrics in Case 2. The proposed methods outperform other baselines in each metric} \label{fig: simulation_case_02}
\end{figure*}

Additionally, this adaptive mechanism improves trajectory prediction accuracy and, subsequently, enhances the responsiveness of intention inference. As shown in Fig.~\ref{fig: case_01_pred_acc_plot} and Fig.~\ref{fig: case_02_pred_acc_plot}, the comparisons between A--B and P--G highlight that accurate $\alpha$ estimation refines the Monte Carlo sampling strategy. For instance, in Case 1, where the target behaves deterministically ($\alpha^{\ast}$ is large), non-adaptive methods G and B provide overly conservative predictions because their fixed $\hat{\alpha}$ is too small. Conversely, in Case 2, where $\hat{\alpha} \gg \alpha^{\ast}$, the non-adaptive methods generate overconfident predictions that lack robustness. Specifically, as shown in Fig.~\ref{fig: simulation_case_02}, the red inference from algorithm G varies a lot between the two consecutive time steps 151 and 152, while the blue inference algorithm P prevents the inference oscillations, providing a more reliable and increasing confidence in the true intention (also see Fig.~\ref{fig: case_02_infer_plot}, time steps 125--170). For the NLL results in Fig~\ref{fig: case_02_pred_nll_plot}, the proposed method maintains the lowest NLL with a substantially slower growth rate, demonstrating that the online adaptation of $\alpha$ enables the predicted probability distribution to remain consistent with the actual target motion.

\begin{figure}[htb]
\centering
\subfloat[Inference and prediction at time step 151]
{\label{fig: case_02_step_151} \includegraphics[width=0.64\linewidth]{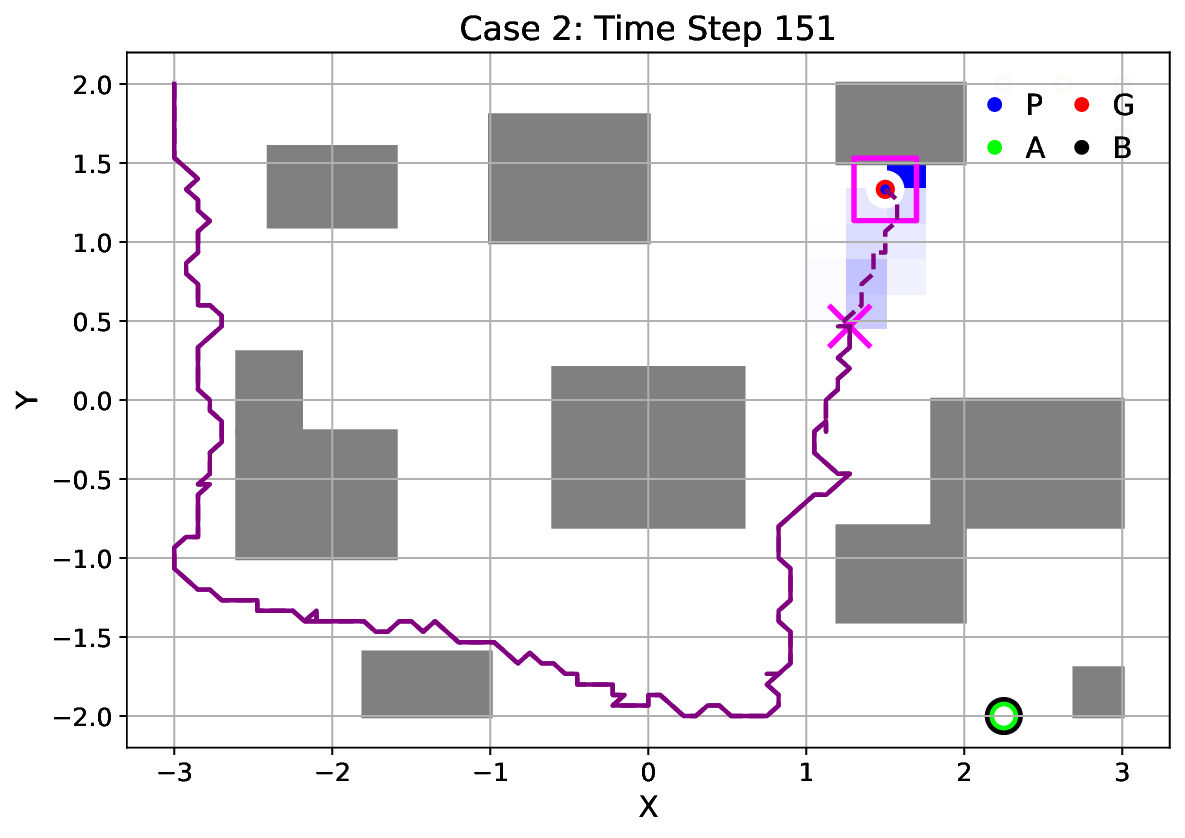}}\\[1ex]
\subfloat[Inference and prediction at time step 152]
{\label{fig: case_02_step_152} \includegraphics[width=0.64\linewidth]{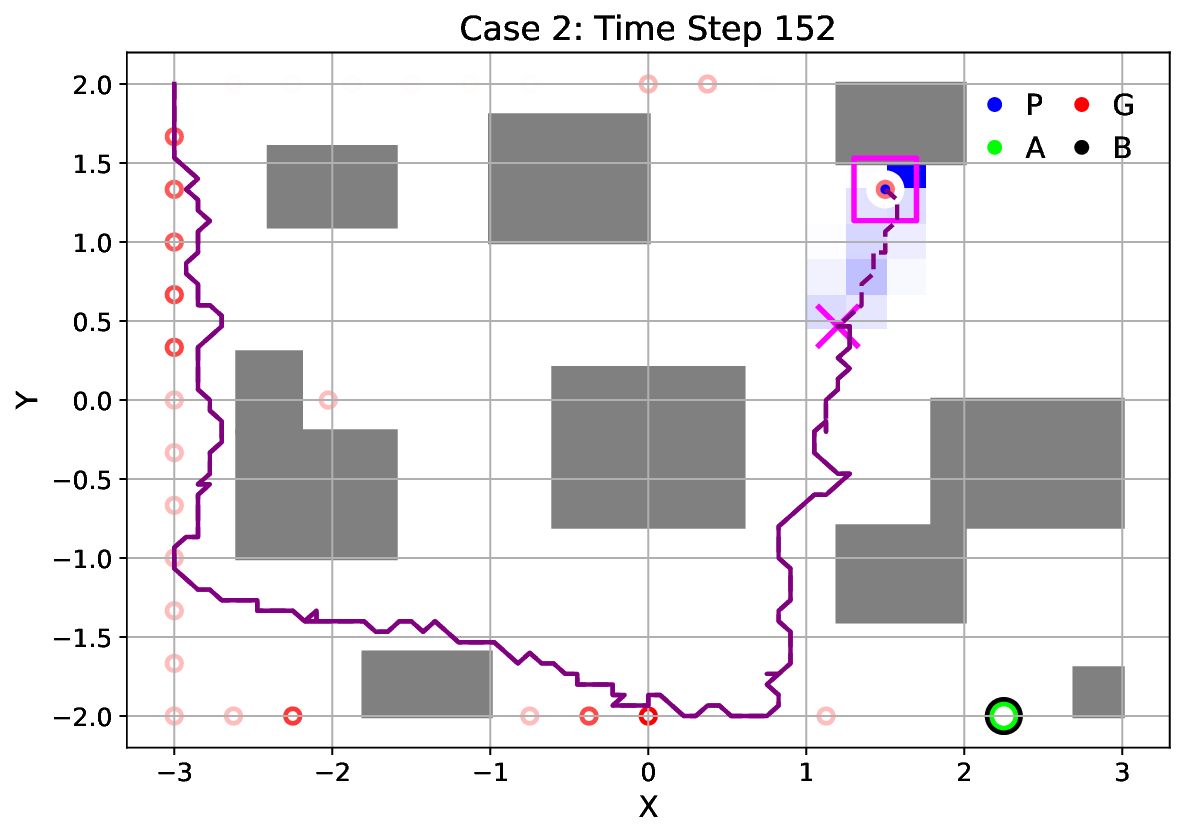}}
\caption{(a) Inference and prediction results for P and inference results for G in Case 2 at $t = 151$. (b) Stable inference P but oscillating inference G on the left half goals in the environment at $t=152$.}\label{fig: simulation_case_02}
\end{figure}

In all, by incorporating the adaptive $\alpha$ estimation module, the proposed algorithm P demonstrates superior performance and robustness through two primary advantages: 1) Enhanced Prediction Accuracy and Probabilistic Robustness: By dynamically updating the intention parameter, the proposed algorithm (P) achieves consistently higher average trajectory prediction accuracy and lower NLL compared to non-adaptive methods (B and G) throughout the evaluation. The adaptive estimation of $\alpha$ allows the model to better align the Monte Carlo samples with the target's actual motion characteristics. 2) Superior Inference Stability: Algorithm P ensures stable intention inference by adapting $\hat{\alpha}$ to the target's specific level of stochasticity.

\begin{figure*}[htb]
\subfloat[Inference probability (higher is better)]
{\label{fig: mc_infer_box_plot} \includegraphics[width=0.32\linewidth]{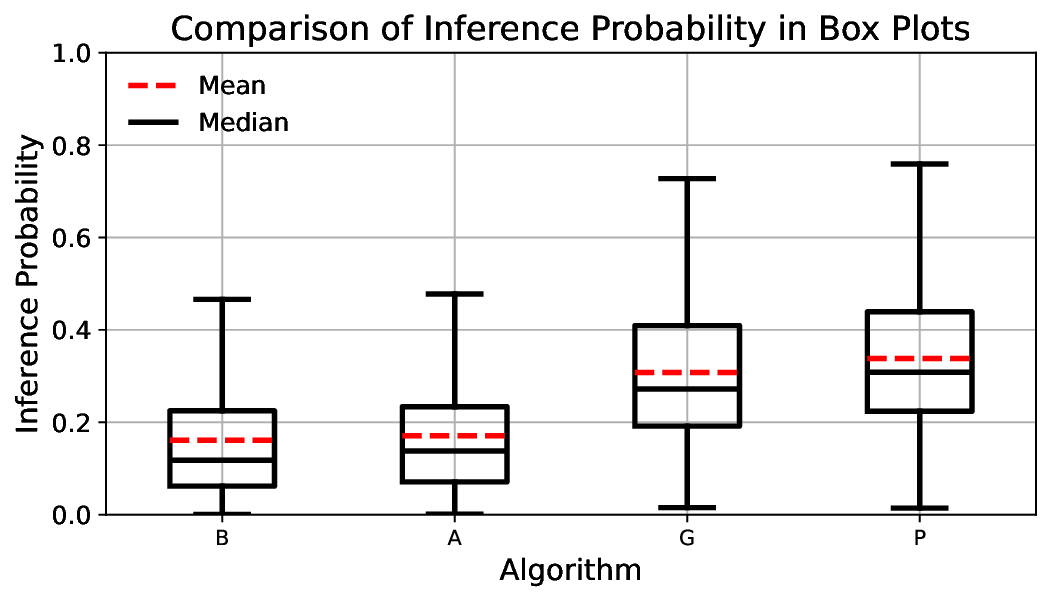}}
\hfill
\subfloat[ACC (higher is better)]
{\label{fig: mc_pred_acc_box_plot} \includegraphics[width=0.32\linewidth]{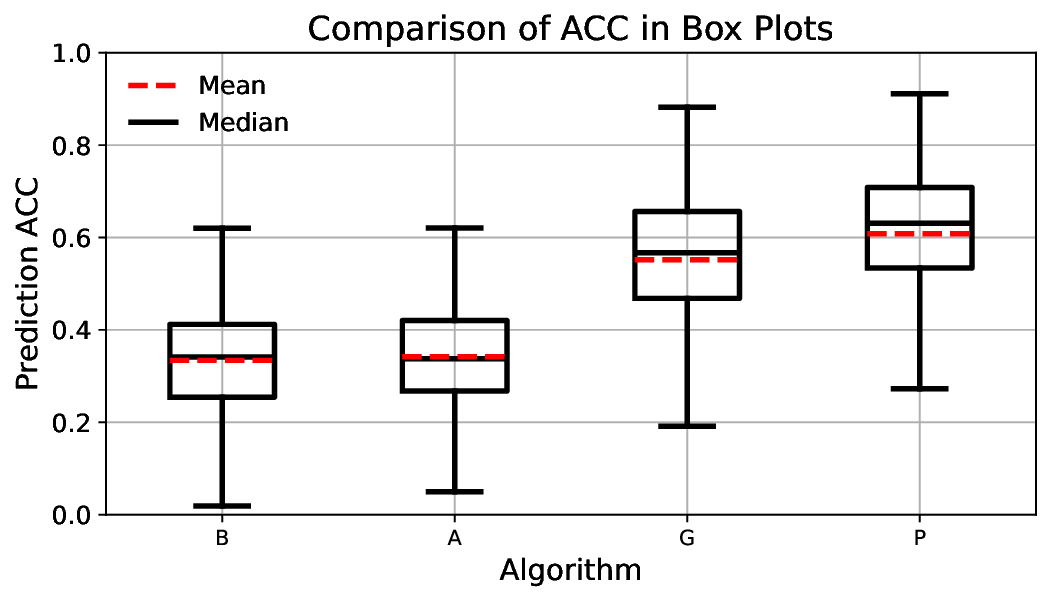}}
\hfill
\subfloat[NLL (lower is better)]
{\label{fig: mc_pred_nll_box_plot} \includegraphics[width=0.32\linewidth]{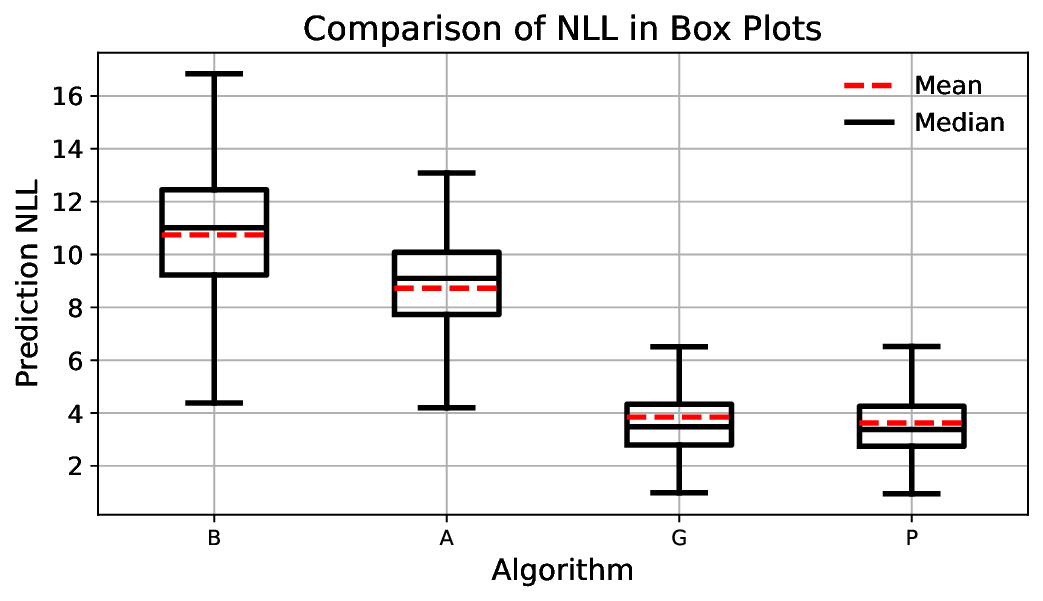}}
\caption{Monte Carlo experiment metrics among different methods} \label{fig: sim_mc}
\end{figure*}

\subsection{Statistical Significance and Scalability}\label{subsec: numerical_results_scale}

\begin{table}[htb]
\centering
\caption{Monte Carlo Comparison}
\label{table: sim_mc}

\renewcommand{\arraystretch}{0.92}
\setlength{\tabcolsep}{2pt}

\begin{tabularx}{0.48\textwidth}{
@{}
>{\centering\arraybackslash}p{0.11\textwidth}
>{\centering\arraybackslash}p{0.14\textwidth}
>{\centering\arraybackslash}p{0.05\textwidth}
>{\centering\arraybackslash}p{0.06\textwidth}
>{\centering\arraybackslash}p{0.08\textwidth}
@{}}
\toprule
Algorithm & Inference Prob. & Acc & NLL & CpT (ms) \\
\midrule
\textbf{B} & 0.17 & 0.33 & 10.74 & \textbf{1.39} \\
\textbf{A} & 0.16 & 0.34 & 8.72 & 1.59 \\
\textbf{G} & 0.31 & 0.55 & 3.85 & 1.49 \\
\textbf{P} & \textbf{0.34} & \textbf{0.61} & \textbf{3.62} & 1.67 \\
\bottomrule
\end{tabularx}

\end{table}

Table~\ref{table: sim_mc} summarizes the results of the 500-trial Monte Carlo experiment. The proposed algorithm (P) achieves the highest average inference probability of $0.34$, demonstrating its superior capacity for intention inference in dynamic environments. Furthermore, the proposed method attains a leading average prediction accuracy of $0.61$, while simultaneously achieving the lowest average negative log-likelihood (NLL) of $3.62$, outperforming all alternative approaches.

Computational performance was evaluated as the average runtime per time step across all generated trajectories. The simulations were conducted on a desktop equipped with an Intel Core i7-13700 CPU and 32~GB of RAM, without utilizing GPU acceleration. While the baseline (B) exhibits the lowest computation time due to its lack of adaptive goal-switching and $\alpha$-estimation modules, algorithm P maintains real-time efficiency while providing significantly enhanced robustness.

To rigorously assess the performance across the four algorithms, we applied a Kruskal–Wallis test \cite{kruskal1952use} to determine if the differences in results are statistically significant, followed by a post-hoc Dunn’s test \cite{dunn1964multiple} for pairwise ranking. The Kruskal–Wallis test is appropriate here as it accommodates distributions with unequal variances. As shown in the box plots in Fig.~\ref{fig: sim_mc}, algorithm P achieves the highest median inference probability, the highest median prediction accuracy as well as the lowest median NLL. The Kruskal–Wallis test successfully rejects the null hypothesis that the four methods share equal means for all three metrics ($p = 0.00 < 0.05$). For inference probability, Dunn’s test confirms that the mean of P is significantly higher than that of B, A, and G, with $p$-values of $0.00$, $0.00$, and $0.0409$, respectively. For prediction accuracy, P also achieves statistically significant improvements over all alternative methods ($p < 0.05$). For NLL, although algorithm P achieves the lowest mean value, Dunn’s test does not identify a statistically significant difference between G and P. This result suggests that the adaptive goal-switching mechanism is the dominant factor influencing the overall probabilistic trajectory distribution, while the adaptive estimation of $\alpha$ primarily refines the concentration and stability of the prediction samples. Consequently, the online adaptation of $\alpha$ provides a stronger improvement in prediction accuracy and inference robustness than in the distribution-level likelihood measured by NLL. These results collectively validate that the proposed method (P) provides the most robust and accurate trajectory prediction under intention and motion uncertainty.

To provide a comprehensive reference for the application of the proposed algorithm in diverse future scenarios, we conduct a systematic scalability test across four critical dimensions: the number of potential goals ($N$), the prediction horizon ($T$), the number of Monte Carlo samples ($M$), and the environment grid resolution. Detailed results are summarized in Table~\ref{table: sim_scalability}.

Within a consistent environment setup, we observe that the computational burden is primarily governed by $T$ and $M$. Specifically, while increasing the number of samples $M$ enhances the prediction accuracy and lowers the NLL by providing a denser approximation of the future state distribution, it also increases the computation time per step. Consequently, users must balance these two factors based on the specific constraints of their platform. Furthermore, we observe that an increase in prediction length $T$ naturally achieves better accuracy and NLL. This occurs because the actual executed trajectory represents only a single realization of the future state distribution; as the prediction horizon expands, the dimensionality of the state space grows, leading to a degradation of these two metrics at the end of the horizon due to the accumulation of stochastic uncertainty.

\begin{table}[htb]
\centering
\caption{Scalability test across different parameters}
\label{table: sim_scalability}
\scriptsize
\setlength{\tabcolsep}{3.0pt}
\begin{tabular}{ccc|ccc|ccc|ccc}
\toprule
& & & \multicolumn{3}{c|}{$61\times41$}
& \multicolumn{3}{c|}{$81\times61$}
& \multicolumn{3}{c}{$101\times81$} \\
\cmidrule(lr){4-6}
\cmidrule(lr){7-9}
\cmidrule(lr){10-12}

$\boldsymbol{N}$ & $\boldsymbol{T}$ & $\boldsymbol{M}$
& \textbf{ACC} & \textbf{NLL} & \textbf{CpT}
& \textbf{ACC} & \textbf{NLL} & \textbf{CpT}
& \textbf{ACC} & \textbf{NLL} & \textbf{CpT} \\
& & & \tiny(\%) & & \tiny(ms)
& \tiny(\%) & & \tiny(ms)
& \tiny(\%) & & \tiny(ms) \\
\midrule

\multirow{9}{*}{50}
& \multirow{3}{*}{10}
& 20
& 72.52 & 2.92 & \textbf{0.26}
& 68.69 & 3.70 & \textbf{0.30}
& 77.53 & 2.13 & \textbf{0.30} \\

& & 100
& 72.88 & 2.37 & 0.45
& 69.26 & 3.12 & 0.55
& 78.07 & 1.64 & 0.57 \\

& & 500
& \textbf{72.97} & \textbf{2.15} & 1.50
& \textbf{69.29} & \textbf{2.88} & 1.80
& \textbf{78.09} & \textbf{1.47} & 1.86 \\

\cmidrule(l){2-12}

& \multirow{3}{*}{20}
& 20
& 55.69 & 5.57 & \textbf{0.31}
& 56.47 & 5.73 & \textbf{0.35}
& 64.41 & 4.11 & \textbf{0.36} \\

& & 100
& 56.10 & 4.76 & 0.65
& 56.90 & 4.87 & 0.81
& 64.99 & 3.40 & 0.85 \\

& & 500
& \textbf{56.22} & \textbf{4.40} & 2.50
& \textbf{56.96} & \textbf{4.50} & 2.93
& \textbf{65.08} & \textbf{3.10} & 3.08 \\

\cmidrule(l){2-12}

& \multirow{3}{*}{30}
& 20
& 45.11 & 7.76 & \textbf{0.33}
& 47.92 & 7.48 & \textbf{0.41}
& 54.87 & 5.88 & \textbf{0.44} \\

& & 100
& 45.56 & 6.79 & 0.79
& 48.36 & 6.51 & 0.99
& 55.49 & 5.02 & 1.07 \\

& & 500
& \textbf{45.67} & \textbf{6.34} & 3.22
& \textbf{48.43} & \textbf{6.03} & 3.90
& \textbf{55.59} & \textbf{4.63} & 4.29 \\

\midrule


\rowcolor{blockgray}
& & 20
& 74.27 & 3.02 & \textbf{0.47}
& 67.80 & 4.47 & \textbf{0.48}
& 74.68 & 2.88 & \textbf{0.52} \\

\rowcolor{blockgray}
& & 100
& 76.74 & 1.93 & 0.70
& 70.08 & 3.28 & 0.76
& 77.14 & 1.65 & 0.81 \\

\rowcolor{blockgray}
& \multirow{-3}{*}{10}
& 500
& \textbf{76.87} & \textbf{1.78} & 1.86
& \textbf{70.11} & \textbf{3.09} & 1.97
& \textbf{77.30} & \textbf{1.51} & 2.06 \\

\cmidrule(l){2-12}

\rowcolor{blockgray}
& & 20
& 59.44 & 5.37 & \textbf{0.52}
& 55.21 & 6.37 & \textbf{0.57}
& 61.68 & 4.94 & \textbf{0.61} \\

\rowcolor{blockgray}
& & 100
& 61.87 & 4.00 & 0.94
& 57.42 & 4.92 & 1.04
& 63.93 & 3.42 & 1.12 \\

\rowcolor{blockgray}
& \multirow{-3}{*}{20}
& 500
& \textbf{62.02} & \textbf{3.76} & 2.96
& \textbf{57.43} & \textbf{4.65} & 3.21
& \textbf{64.06} & \textbf{3.13} & 3.42 \\

\cmidrule(l){2-12}

\rowcolor{blockgray}
& & 20
& 49.41 & 7.35 & \textbf{0.56}
& 46.41 & 8.05 & \textbf{0.64}
& 52.29 & 6.72 & \textbf{0.68} \\

\rowcolor{blockgray}
& & 100
& 51.62 & 5.86 & 1.12
& 48.37 & 6.47 & 1.25
& 54.28 & 5.05 & 1.35 \\

\rowcolor{blockgray}
\multirow{-9}{*}{100}
& \multirow{-3}{*}{30}
& 500
& \textbf{51.79} & \textbf{5.54} & 3.73
& \textbf{48.46} & \textbf{6.13} & 4.16
& \textbf{54.35} & \textbf{4.67} & 4.60 \\

\midrule

\multirow{9}{*}{150}
& \multirow{3}{*}{10}
& 20
& 65.22 & 5.07 & \textbf{0.73}
& 64.88 & 5.21 & \textbf{0.76}
& 71.55 & 3.70 & \textbf{0.81} \\

& & 100
& 74.25 & 1.99 & 1.03
& 69.19 & 3.28 & 1.05
& 75.87 & 1.75 & 1.12 \\

& & 500
& \textbf{74.31} & \textbf{1.79} & 2.14
& \textbf{69.26} & \textbf{3.03} & 2.22
& \textbf{75.98} & \textbf{1.55} & 2.34 \\

\cmidrule(l){2-12}

& \multirow{3}{*}{20}
& 20
& 50.38 & 7.55 & \textbf{0.79}
& 52.07 & 7.23 & \textbf{0.83}
& 58.63 & 5.84 & \textbf{0.91} \\

& & 100
& 58.00 & 4.12 & 1.21
& 56.07 & 4.99 & 1.27
& 62.27 & 3.54 & 1.44 \\

& & 500
& \textbf{58.09} & \textbf{3.80} & 3.18
& \textbf{56.24} & \textbf{4.60} & 3.45
& \textbf{62.37} & \textbf{3.20} & 3.73 \\

\cmidrule(l){2-12}

& \multirow{3}{*}{30}
& 20
& 40.67 & 9.55 & \textbf{0.81}
& 43.36 & 8.94 & \textbf{0.90}
& 49.48 & 7.63 & \textbf{0.98} \\

& & 100
& 47.02 & 6.07 & 1.36
& 47.03 & 6.62 & 1.51
& 52.51 & 5.16 & 1.63 \\

& & 500
& \textbf{47.20} & \textbf{5.64} & 3.90
& \textbf{47.23} & \textbf{6.11} & 4.39
& \textbf{52.63} & \textbf{4.71} & 4.92 \\

\bottomrule
\end{tabular}
\vspace{1pt}
\parbox{\columnwidth}{
Results are compared for each fixed combination of the number of candidate goals $N$, prediction horizon $T$, number of Monte Carlo samples $M$, and grid resolution. Higher \textbf{ACC}, lower \textbf{NLL} and \textbf{CpT} indicate better performance.
}
\end{table}

The increase in the number of candidate goals $N$ and the grid resolution also contributes to higher computation time. The impact of $N$ is a direct result of the expanded Bayesian update loop required for each candidate goal, while the grid size affects the dimensions of the stored matrix for the shortest-path attribute $\delta(\cdot,\cdot)$. We also note that the improvement in prediction accuracy and NLL gained from a larger $M$ is more pronounced as the grid size increases. Also, our results show that this gain becomes marginal beyond a certain sample size. Specifically, in a $101 \times 81$ map, increasing $M$ from 100 to 500 yielded negligible performance improvements, which drives our decision to upper-bound $M$ at 500 for the simulations and hardware experiments in this work to preserve high update frequencies. Even at the highest evaluated complexity ($N=150$ and a $101 \times 81$ grid), the algorithm maintains a mean computation time of approximately $\SI{4.92}{ms}$, confirming its capability for real-time operation at approximately $\SI{203}{Hz}$.

\section{Experimental Evaluation}\label{sec: experiment}
In Section~\ref{sec: comparison learning}, we compare the proposed Bayesian inference algorithm with several state-of-the-art learning-based trajectory prediction methods on the ETH/UCY benchmark datasets. The results demonstrate that the proposed approach achieves improved prediction robustness while maintaining competitive prediction accuracy. In Section~\ref{sec: harward experiment}, we further validate the proposed algorithm through real-world hardware experiments on quadruped and quadrotor platforms, demonstrating its real-time inference and trajectory prediction capability.
\subsection{Evaluation on Public Datasets}\label{sec: comparison learning}
\subsubsection{Dataset Description and Preprocessing}

The proposed method is evaluated on the ETH~\cite{pellegrini2010improving} and UCY~\cite{lerner2007crowds} pedestrian trajectory datasets, which are widely used benchmarks for trajectory prediction research. The ETH dataset consists of two scenes, namely ETH and HOTEL, while the UCY dataset contains three scenes: UNIV, ZARA1, and ZARA2. Each dataset provides pedestrian trajectories coordinates in meters together with annotated obstacle regions within the environment.

To construct the environment cost matrix required by the proposed algorithm, the obstacle annotations are first transformed into the same coordinate frame as the pedestrian trajectories using the provided homography matrices. The environment is then discretized into a square grid representation, and the traversal cost between neighboring grid cells is computed using the A* algorithm~\cite{devaurs2015optimal}. Pedestrian trajectories are linearly interpolated between recorded waypoints and subsequently mapped onto the discretized grid environment.

Unlike many learning-based baseline methods that utilize a fixed observation horizon of 3.2 seconds (8 frames) to predict the subsequent 4.8 seconds (12 frames), the proposed algorithm is not restricted to a fixed observation window and can naturally incorporate variable-length trajectory observations. Since the proposed Bayesian inference algorithm incrementally updates belief states over time, incorporating the complete observation history improves intention estimation consistency.

To establish a fair comparison with prior learning-based approaches, the prediction horizon of the proposed algorithm is aligned with the standard 4.8 second prediction interval used in ETH/UCY evaluations. Since the proposed algorithm operates on a discretized grid representation rather than directly in continuous time, a mapping between prediction timesteps and physical time duration is required. Instead of assuming a constant-velocity dynamic model for all pedestrians, we estimate the average pedestrian velocity $V_{ave}$ for each dataset and determine the corresponding prediction horizon as $T = \frac{4.8d(1+\sqrt{2})}{2V_{ave}}$, where $d$ denotes the grid resolution. The grid resolution $d=\SI{0.2}{m}$ was selected to balance computational efficiency and trajectory representation accuracy. The term $\frac{d(1+\sqrt{2})}{2}$ represents the average Euclidean traversal distance between neighboring grid cells in the square grid environment. Table~\ref{table: dataset_specs} summarizes the environment dimensions, grid resolutions, average pedestrian velocities, and corresponding prediction horizons used for each dataset.
\begin{table}[htbp]
\centering
\caption{Environment Configuration and Prediction Parameters}
\label{table: dataset_specs}
\begin{tabular}{@{}lccccc@{}}
\toprule
 & \textbf{ETH} & \textbf{HOTEL} & \textbf{UNIV} & \textbf{ZARA1} & \textbf{ZARA2} \\ \midrule
Map size (m) & 24$\times$18 & 8$\times$15 & 15$\times$14 & 15$\times$15 & 15$\times$15 \\
Grid resolution (m) & \multicolumn{5}{c}{0.2} \\
Prediction horizon & 28 & 23 & 14 & 22 & 22 \\
\bottomrule
\end{tabular}
\end{table}

\begin{table*}[htb]
\centering
\caption{Trajectory prediction performance on ETH/UCY datasets. Lower is better.}
\label{table: comparison_results}
\setlength{\tabcolsep}{4pt}
\renewcommand{\arraystretch}{1.15}
\small

\begin{tabular}{llcccccccccc}
\toprule

& &
\multicolumn{2}{c}{ETH}
& \multicolumn{2}{c}{HOTEL}
& \multicolumn{2}{c}{UNIV}
& \multicolumn{2}{c}{ZARA1}
& \multicolumn{2}{c}{ZARA2} \\

\cmidrule(lr){3-4}
\cmidrule(lr){5-6}
\cmidrule(lr){7-8}
\cmidrule(lr){9-10}
\cmidrule(lr){11-12}

Method & Eval.
& ADE & FDE
& ADE & FDE
& ADE & FDE
& ADE & FDE
& ADE & FDE \\

\midrule
S-GAN \cite{gupta2018social}
& BoN
& 0.87 & 1.62
& 0.67 & 1.37
& 0.76 & 1.52
& 0.35 & 0.68
& 0.42 & 0.84 \\

S-LSTM \cite{alahi2016social}
& BoN
& 1.09 & 2.35
& 0.79 & 1.67
& 0.67 & 1.40
& 0.47 & 1.00
& 0.56 & 1.17 \\

Social-STGCNN \cite{mohamed2020social}
& BoN
& 0.64 & 1.11
& 0.49 & 0.85
& 0.44 & 0.79
& 0.34 & 0.53
& 0.30 & 0.48 \\

Flow-Chain \cite{maeda2023fast}
& BoN
& 0.55 & 0.99
& 0.20 & 0.35
& 0.29 & 0.54
& 0.22 & 0.40
& 0.20 & 0.34 \\

Social-implicit \cite{mohamed2022social}
& BoN
& 0.66 & 1.44
& 0.20 & 0.36
& 0.31 & 0.60
& 0.25 & 0.50
& 0.22 & 0.43 \\

SingularTrajectory \cite{bae2024singulartrajectory}
& BoN
& 0.35 & 0.42
& 0.13 & 0.19
& 0.25 & 0.44
& 0.19 & 0.32
& 0.15 & 0.25 \\

\midrule

Flow-Chain \cite{maeda2023fast}
& Mean-N
& \stdcell{1.24}{0.74} & \stdcell{2.43}{1.43}
& \stdcell{0.59}{0.25} & \stdcell{1.12}{0.50}
& \stdcell{0.70}{0.27} & \stdcell{1.43}{0.54}
& \stdcell{0.59}{0.24} & \stdcell{1.19}{0.49}
& \stdcell{0.57}{0.24} & \stdcell{1.14}{0.49} \\

Social-implicit \cite{mohamed2022social}
& Mean-N
& \stdcell{1.07}{0.23} & \stdcell{2.36}{0.48}
& \stdcell{0.78}{0.37} & \stdcell{1.48}{0.71}
& \stdcell{0.73}{0.25} & \stdcell{1.49}{0.51}
& \stdcell{0.56}{0.18} & \stdcell{1.14}{0.36}
& \stdcell{0.54}{0.19} & \stdcell{1.19}{0.4} \\

SingularTrajectory \cite{bae2024singulartrajectory}
& Mean-N
& \stdcell{1.47}{1.27} & \stdcell{2.78}{1.97}
& \stdcell{0.89}{0.60} & \stdcell{1.89}{1.27}
& \stdcell{1.13}{0.64} & \stdcell{2.37}{1.34}
& \stdcell{0.92}{0.57} & \stdcell{1.97}{1.22}
& \stdcell{1.09}{0.62} & \stdcell{2.24}{1.30} \\

\textbf{Ours}
& Mean-N
& \stdcell{\textbf{0.46}}{\textbf{0.011}}
& \stdcell{\textbf{0.81}}{\textbf{0.013}}

& \stdcell{\textbf{0.34}}{\textbf{0.008}}
& \stdcell{\textbf{0.56}}{\textbf{0.010}}

& \stdcell{\textbf{0.40}}{\textbf{0.009}}
& \stdcell{\textbf{0.75}}{\textbf{0.013}}

& \stdcell{\textbf{0.37}}{\textbf{0.008}}
& \stdcell{\textbf{0.67}}{\textbf{0.009}}

& \stdcell{\textbf{0.42}}{\textbf{0.009}}
& \stdcell{\textbf{0.75}}{\textbf{0.010}} \\

\bottomrule
\end{tabular}
\vspace{1mm}
\begin{minipage}{\linewidth}
\footnotesize
\raggedright
BoN denotes the best-of-$N$ evaluation method commonly used in prior stochastic prediction methods.
Mean-20 reports repeated-trial average performance over $N=20$ evaluations, reported as mean $\pm$ standard deviation.
\end{minipage}
\end{table*}

\subsubsection{Comparison with Baselines}
The evaluation metrics are chosen as ADE and FDE, which are widely used in trajectory prediction research~\cite{mohamed2022social, maeda2023fast}. Recent stochastic trajectory prediction methods are typically evaluated using a best-of-$N$ (BoN) strategy, where the model generates $N$ trajectory samples (commonly $N=20$), and the sample with the minimum prediction error with respect to the ground truth is selected for evaluation. Such evaluation strategy has also been discussed and criticized in prior work~\cite{mohamed2022social}, since it measures the best achievable prediction among multiple samples rather than the expected prediction quality across repeated evaluations.

We argue that such an evaluation strategy may not fully reflect the practical requirements of following robotics applications. In many real-world scenarios, trajectory predictions are further utilized for tasks such as motion planning, risk assessment, or threat evaluation, where prediction consistency and reliability are critical. In these applications, a quantitative decision must be made based on the predicted trajectories, rather than selecting the single closest prediction to the ground truth after observing future outcomes. Therefore, in addition to reporting the published BoN results, we further evaluate representative stochastic prediction methods using a repeated-trial average evaluation strategy, denoted as Mean-N, where ADE/FDE are averaged over N evaluations. $N=20$ in our experiments. Furthermore, we report the corresponding standard deviations to characterize prediction stability across repeated trials.

Table~\ref{table: comparison_results} compares the proposed method with several representative trajectory prediction approaches, including S-GAN~\cite{gupta2018social}, S-LSTM~\cite{alahi2016social}, Social-STGCNN~\cite{mohamed2020social}, Flow-Chain~\cite{maeda2023fast}, Social-Implicit~\cite{mohamed2022social}, and SingularTrajectory~\cite{bae2024singulartrajectory}. Existing algorithms, especially stochastic models, are typically evaluated with the BoN strategy, which reports the minimum prediction error among multiple generated trajectory samples. In this case, methods such as SingularTrajectory and Flow-Chain achieve strong ADE/FDE performance due to their ability to generate diverse trajectory hypotheses.

However, under Mean-N evaluation, these stochastic generative approaches experience substantial performance degradation accompanied by large variances. For example, SingularTrajectory achieves an ETH ADE/FDE of $0.35/0.42$ under BoN evaluation, but degrades to $1.47 \pm 1.27$ and $2.78 \pm 1.97$ under Mean-20 evaluation. Similar trends are observed for Flow-Chain and Social-Implicit across all datasets, indicating that only a subset of generated trajectories achieves low prediction error while the overall prediction quality across repeated evaluations remains highly variable.

In contrast, the proposed Bayesian inference algorithm maintains highly stable performance across repeated evaluations, consistently achieving lower variance on all datasets. For example, the proposed method achieves an ETH ADE of $0.46 \pm 0.007$, while maintaining standard deviations on the order of $10^{-2}$ across all scenarios. Unlike stochastic generative approaches that rely on best sample selection, the proposed method produces reliable and consistent trajectory predictions with substantially reduced sensitivity to frequent evaluations. This property is particularly desirable for the following robotics and autonomous navigation applications, where stable prediction behavior is often more important than a "best case" performance without knowing the ground truth in advance.

\subsection{Hardware Experiments}\label{sec: harward experiment}
The proposed algorithm is demonstrated on both a quadrotor drone and a quadrupedal robot. The target’s intention is defined as its current goal, the location a robot seeks to reach.
\subsubsection{Experiment Setup}

The physical testing environment measures 4.8 m in length and 3.6 m in width. To maintain consistency with our numerical evaluations, we employ two distinct obstacle configurations, both of which are discretized into an $81 \times 61$ grid. This resolution provides a balance between spatial fidelity and the high-frequency update requirements of our algorithm. We conduct a total of eight experiments, four on the quadrotor drone and four on the quadrupedal robot. In each trial, the robotic platform starts from a different initial position and attempts to follow the shortest path to a goal position manually manipulated during the experiment.
This human operation may not follow exactly the assumption on the target movement made in Section~\ref{sec: problem_formulation}.
Nonetheless, the following results demonstrate that the proposed algorithm remains effective in intention inference and trajectory prediction, producing accurate goal estimation and reliable estimation $\hat{\alpha}$. The robotic platform changes its intended goal multiple times before ultimately reaching the final goal, which is unknown to the inference algorithm. A motion capture system is used to observe the robot's position in real time. This positional data is streamed to the proposed algorithm, which then estimates the current goal and predicts the future trajectory accordingly.

\subsubsection{Hardware Results}
\begin{figure*}
\subfloat[Before the first intention change in quadrotor experiment]
{\label{fig: exp_quad_down_01} \includegraphics[width=0.3\linewidth]{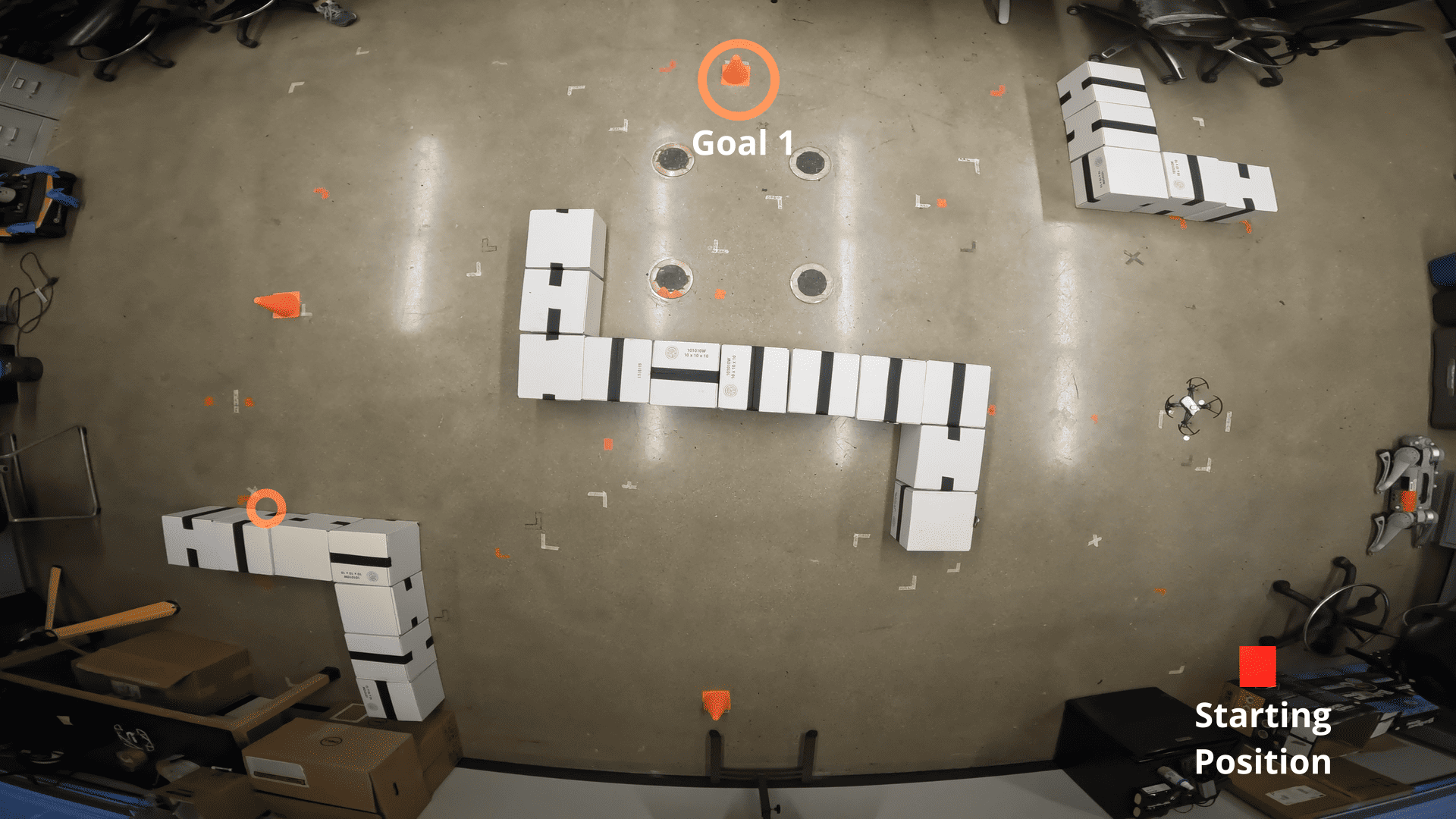}}
\hfill
\subfloat[Before the second intention change in quadrotor experiment]
{\label{fig: exp_quad_down_02} \includegraphics[width=0.3\linewidth]{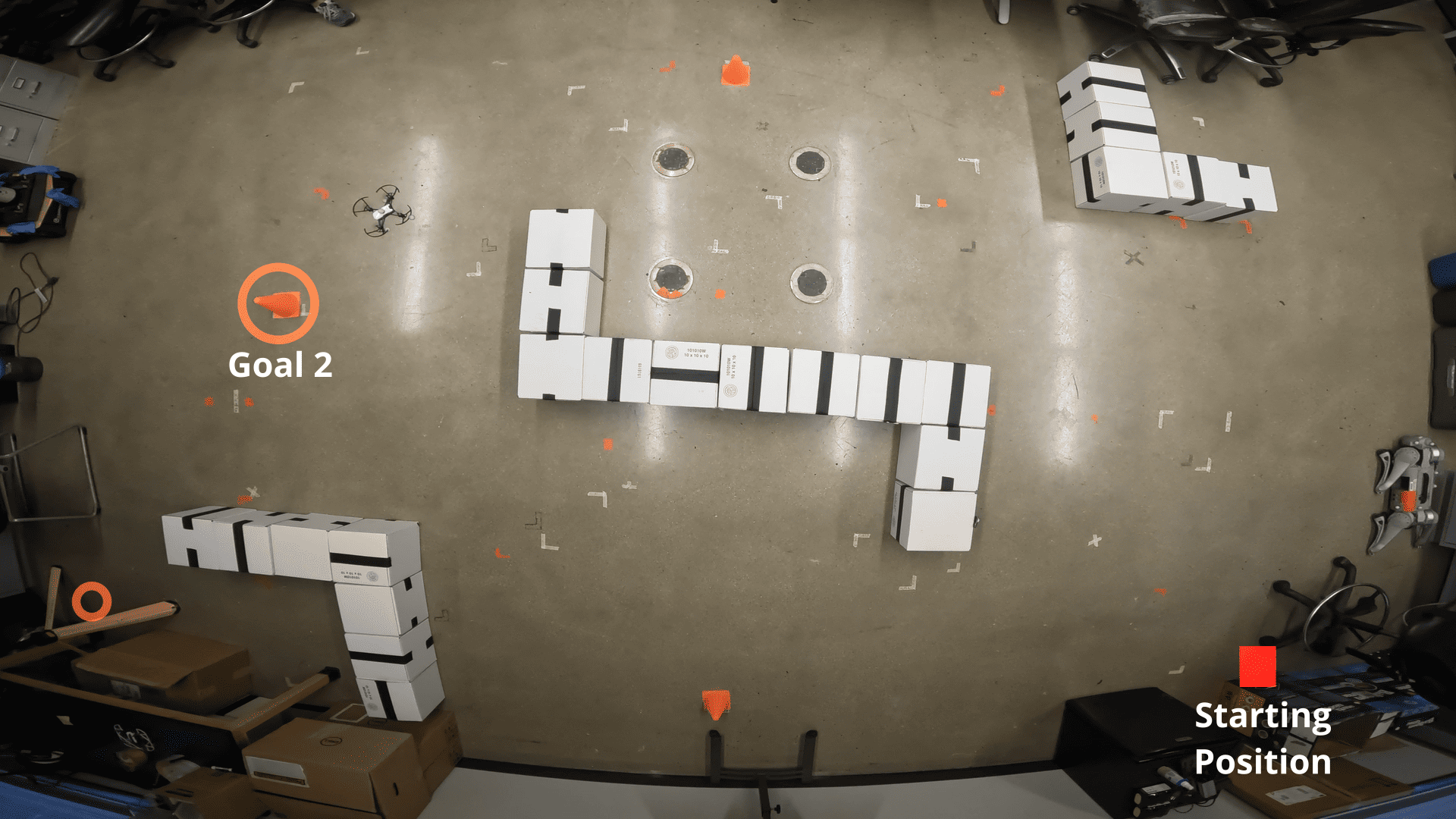}}
\hfill
\subfloat[Before the reaching the terminal goal in quadrotor experiment]
{\label{fig: exp_quad_down_03} \includegraphics[width=0.3\linewidth]{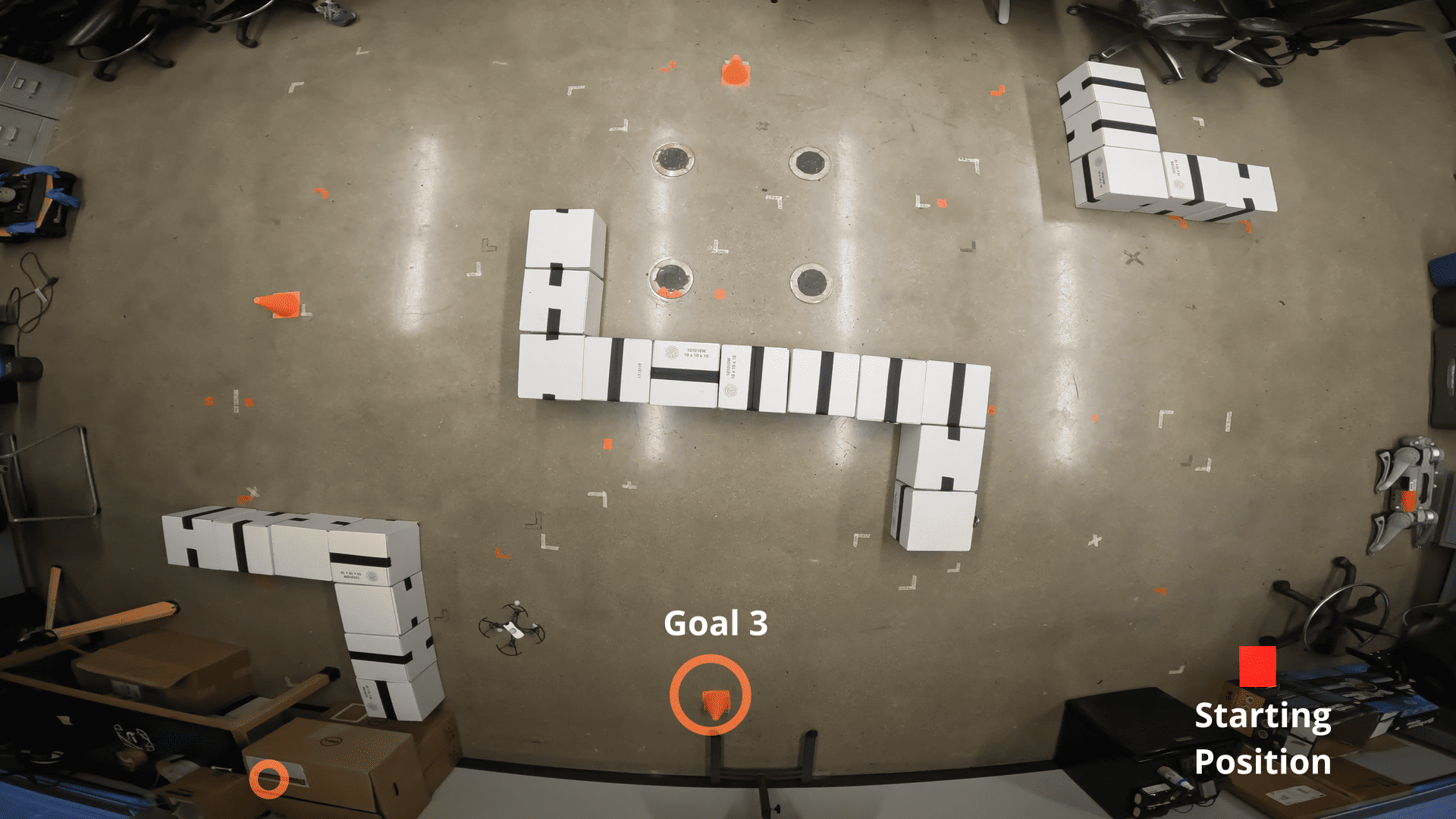}}
\hfill
\subfloat[Before the first intention change in quadrotor experiment]
{\label{fig: exp_quad_alg_01} \includegraphics[width=0.3\linewidth]{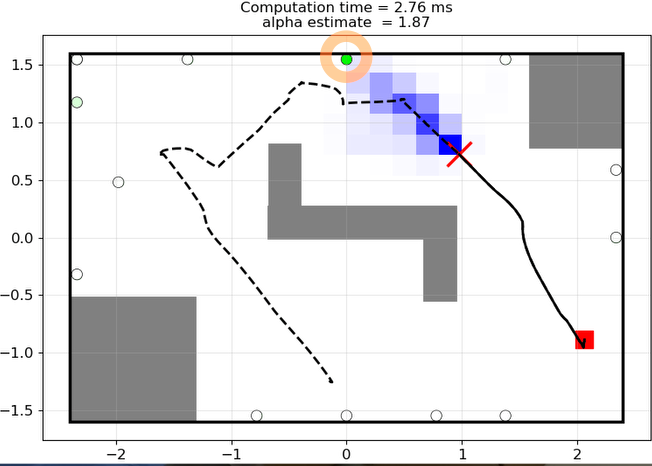}}
\hfill
\subfloat[Before the second intention change in quadrotor experiment]
{\label{fig: exp_quad_alg_02} \includegraphics[width=0.3\linewidth]{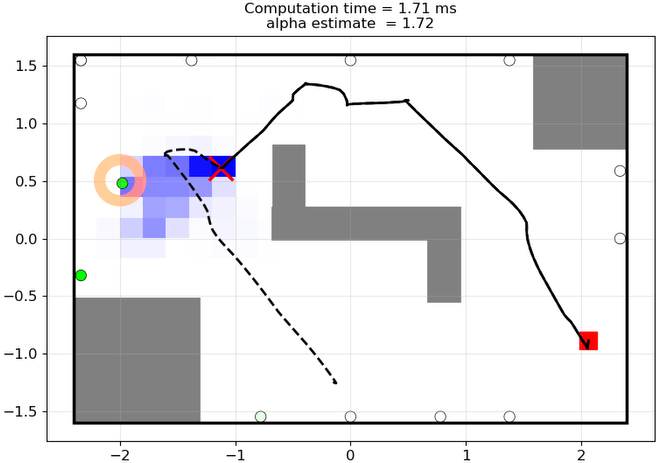}}
\hfill
\subfloat[Before the reaching the terminal goal in quadrotor experiment]
{\label{fig: exp_quad_alg_03} \includegraphics[width=0.3\linewidth]{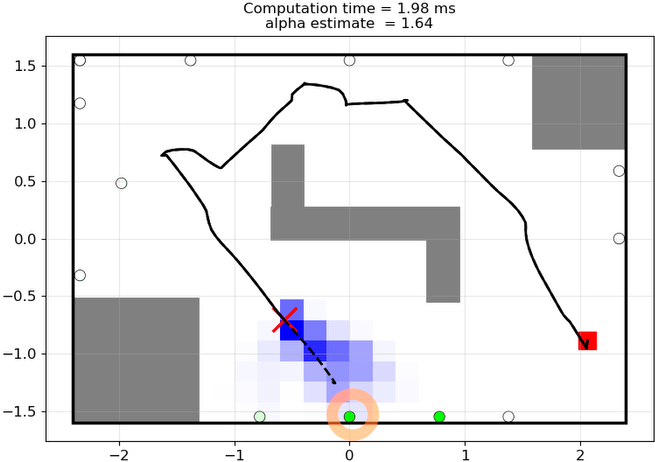}}
\hfill
\subfloat[Before the first intention change in dog experiment]
{\label{fig: exp_dog_down_01} \includegraphics[width=0.3\linewidth]{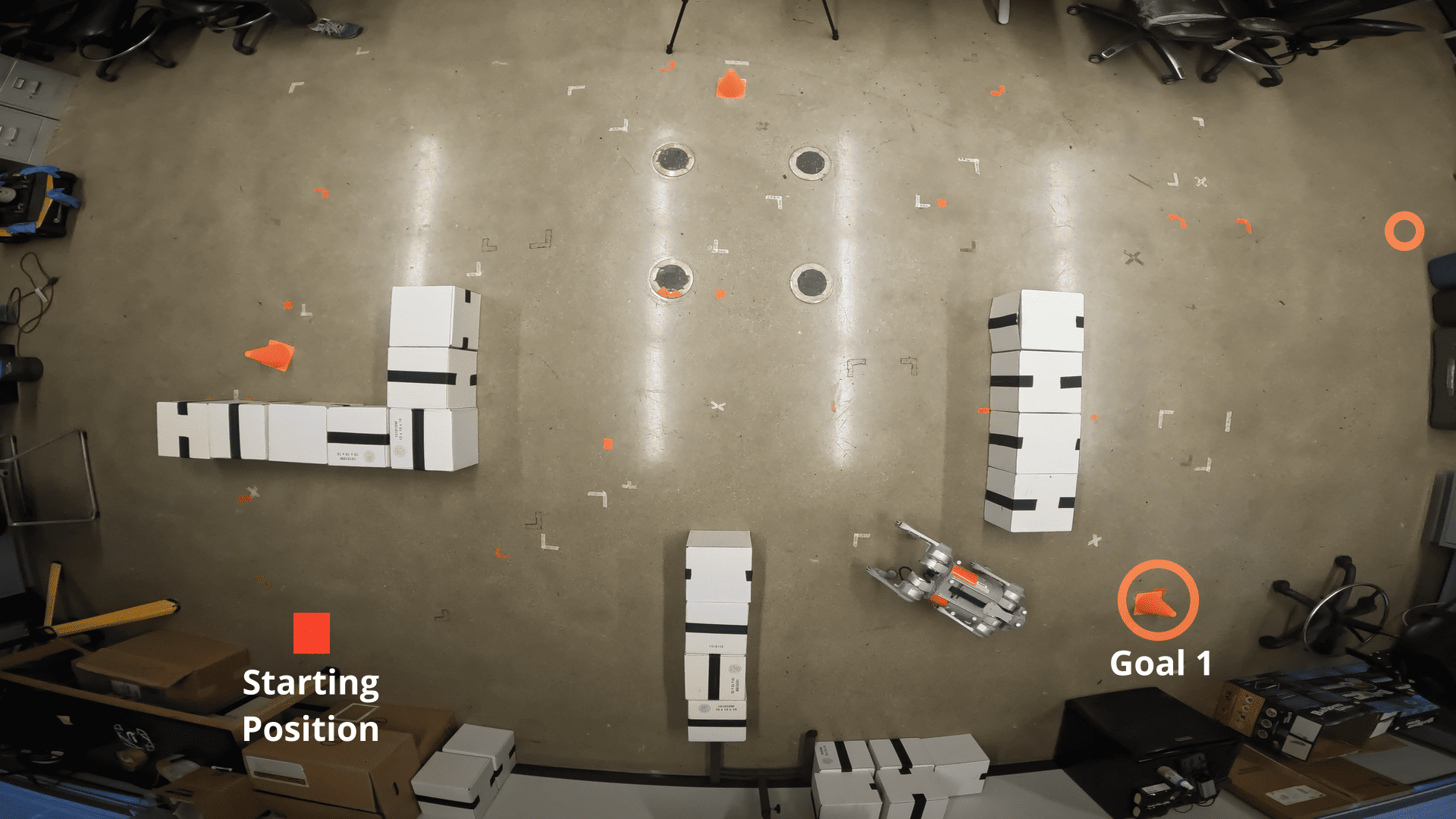}}
\hfill
\subfloat[Before the second intention change in dog experiment]
{\label{fig: exp_dog_down_02} \includegraphics[width=0.3\linewidth]{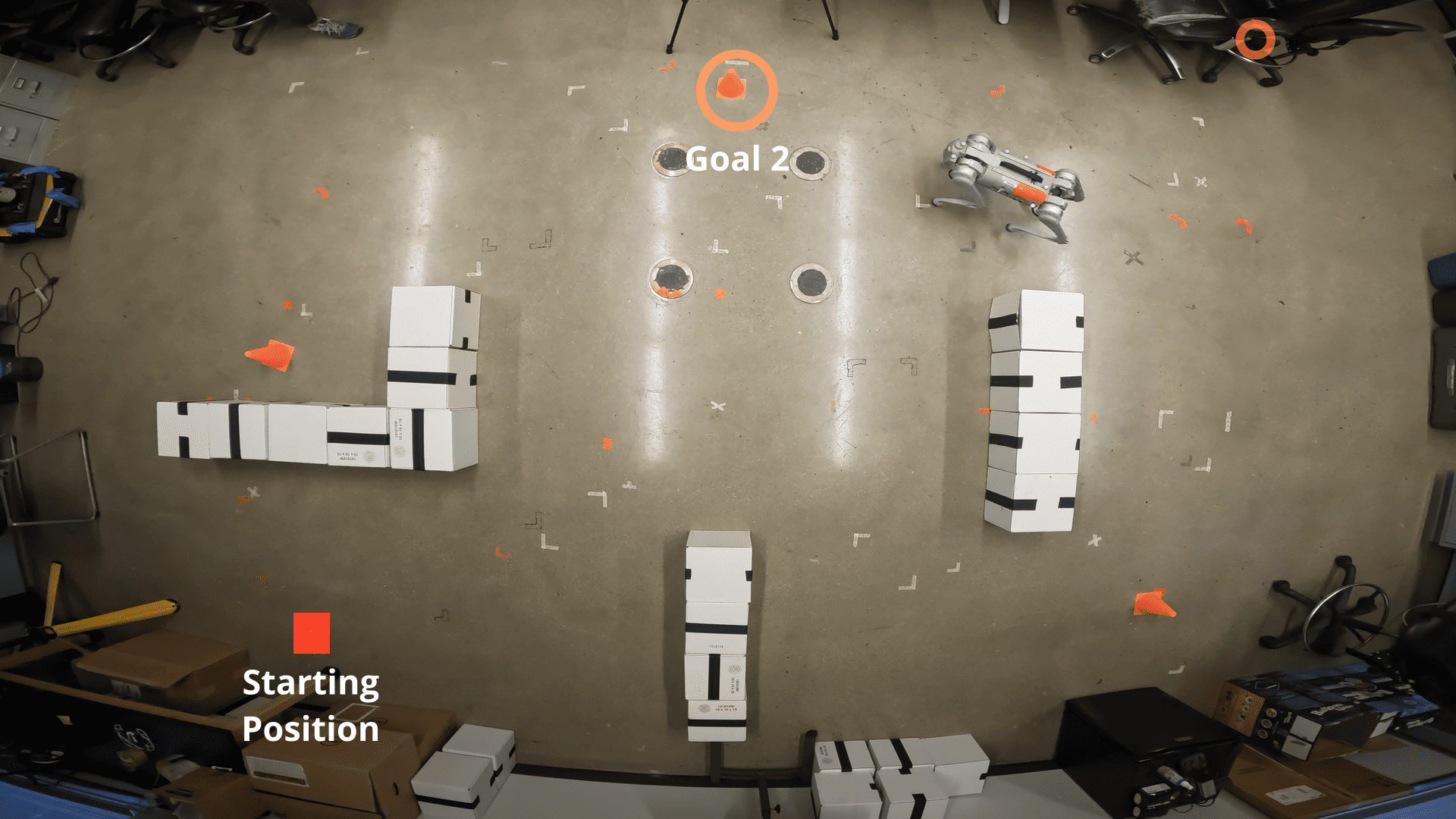}}
\hfill
\subfloat[Before the reaching the terminal goal in dog experiment]
{\label{fig: exp_dog_down_03} \includegraphics[width=0.3\linewidth]{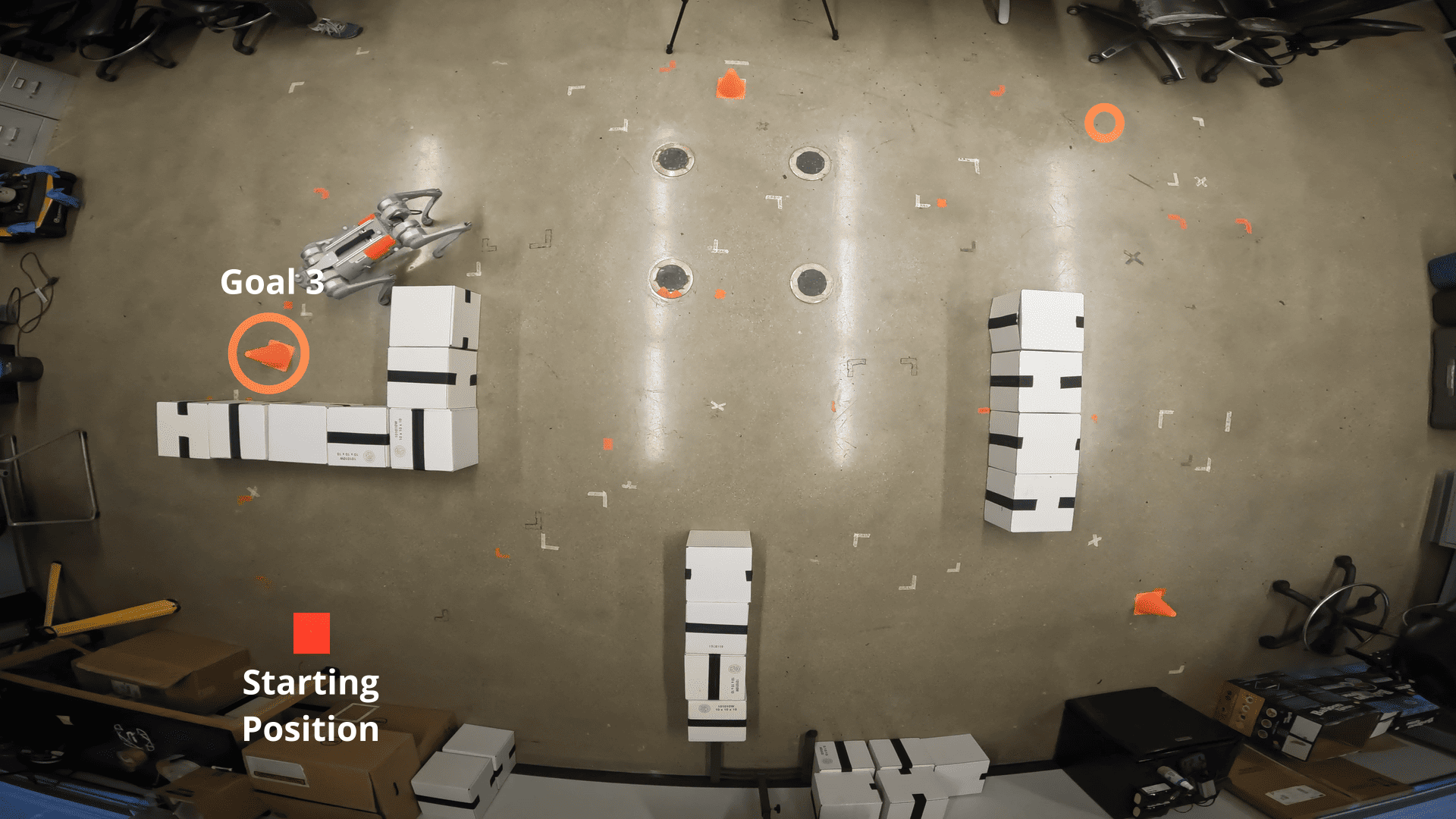}}
\hfill
\subfloat[Before the first intention change in dog experiment]
{\label{fig: exp_dog_alg_01} \includegraphics[width=0.3\linewidth]{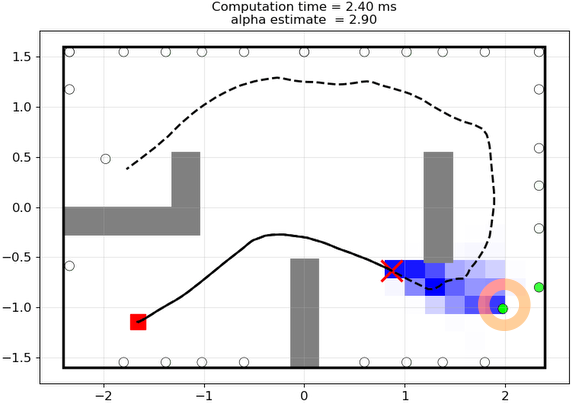}}
\hfill
\subfloat[Before the second intention change in dog experiment]
{\label{fig: exp_dog_alg_02} \includegraphics[width=0.3\linewidth]{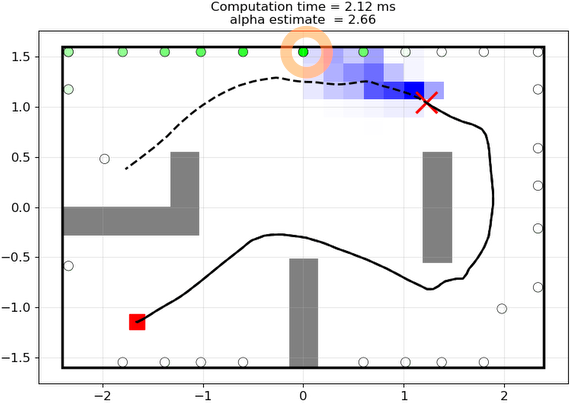}}
\hfill
\subfloat[Before the reaching the terminal goal in dog experiment]
{\label{fig: exp_dog_alg_03} \includegraphics[width=0.3\linewidth]{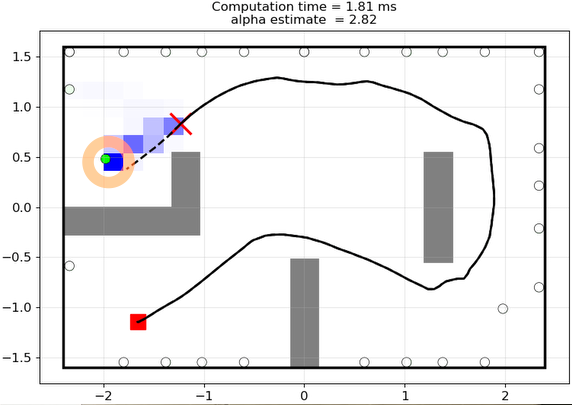}}
\caption{The top-down-view of the experiment and the corresponding real-time inference and prediction for the quadrotor and dog. Each subplot illustrates the target's current goal (orange cone in top-down view and circle in plots) at the current time. The red square marks the target’s initial position, and the red cross indicates its current position. The dashed black line shows the reference trajectory that the quadrotor followed in reality, which is unknown to the inference algorithm. Circles represent candidate goal positions. The level of green shading in each circle reflects the inferred intention: the darker the green, the higher the estimated probability that the corresponding candidate is the target's current intended goal. Trajectory predictions are visualized through blue bins; the intensity of the color correlates with the normalized visitation frequency of Monte Carlo samples, providing a direct measure of prediction confidence and spatial uncertainty.} \label{fig:exp_quadrotor_dog}
\end{figure*}

The experimental results are illustrated in Fig.~\ref{fig:exp_quadrotor_dog}. Fig.\ref{fig: exp_quad_down_01}–\ref{fig: exp_quad_alg_03} shows a quadrotor case where the target sequentially targets three distinct goals. As shown in Fig~\ref{fig: exp_quad_alg_01}, before the first shift, the algorithm maintains a concentrated belief on Goal 1 (top-middle left). Despite the quadrotor's maneuvers based on the changed intention, the inference module rapidly identifies the switch to Goal 2 (bottom left) and finally Goal 3 (top-middle right), with the inferred goal (darkest green circle) consistently aligning with the actual target (orange cone). The blue bins in Fig~\ref{fig: exp_quad_alg_03} aligned well with the actual trajectory. This is driven by the online estimation of a proper intention parameter $\alpha$, reflecting the nature of the quadrotor's path-following behavior. Fig.~\ref{fig: exp_dog_down_01}–\ref{fig: exp_dog_alg_03} display the quadrupedal robot experiment. Although the quadruped moves faster than the quadrotor in our experiments, the algorithm maintains reliable prediction. The quadruped follows a nontrivial trajectory to avoid obstacles while switching from the bottom-left to the top-middle right. Even under rapid motion, the joint update of $\alpha$ and $\theta_k$ prevents inference oscillations, providing reliable future state distributions as the robot navigates toward the terminal goal.

The algorithm is implemented in Python and executed on a desktop with an Intel Core i7-13700 CPU and 32 GB of RAM. No GPU acceleration was used in any of the experiments. As discussed in Remark \ref{remark: parallel_compute}, the algorithm supports parallel implementation, further highlighting its potential for time-constrained applications. Also, the algorithm's computation performance can be further enhanced through a C++ implementation combined with coding optimization. Under the above hardware condition, the computational performance is favorable: the maximum computation time per time step is less than 3.2 ms, and the mean runtime across all eight experiments is 1.83 ms. This supports real-time operation at approximately 547 Hz. This real-time capability is also demonstrated in the supplementary video, where the computation time per step is visualized. These results confirm that the proposed algorithm delivers robust, real-time intention inference and trajectory prediction across multiple robotic platforms. Further details are available in the supplementary video.

\section{Conclusion}\label{sec: conclusion}
This paper presents an adaptive Bayesian algorithm for intention inference and trajectory prediction in scenarios where both the target’s goal and trajectory behavior are unknown and may change over time. By jointly estimating the goal intention and the latent intention parameter $\alpha$, the proposed algorithm captures both the intention change and the unknown kinematics.
Through numerical and real-world experiments, we demonstrated that the proposed method consistently outperforms baseline and partially adaptive methods in terms of both inference accuracy and prediction robustness. Monte Carlo evaluation and scalability test further confirmed that the method scales well and remains reliable across diverse scenarios. 
omparison with state-of-the-art learning-based methods on the public ETH/UCY dataset demonstrates that the proposed algorithm delivers stable and accurate performance.
The algorithm operates in real-time at approximately \SI{547}{Hz} without requiring extensive training on any dataset, making it broadly applicable to a variety of real-world platforms where communication or cooperation from the target is unavailable. 

\bibliographystyle{IEEEtran}
\bibliography{reference}

\begin{thebibliography}{10}
\providecommand{\url}[1]{#1}
\csname url@samestyle\endcsname
\providecommand{\newblock}{\relax}
\providecommand{\bibinfo}[2]{#2}
\providecommand{\BIBentrySTDinterwordspacing}{\spaceskip=0pt\relax}
\providecommand{\BIBentryALTinterwordstretchfactor}{4}
\providecommand{\BIBentryALTinterwordspacing}{\spaceskip=\fontdimen2\font plus
\BIBentryALTinterwordstretchfactor\fontdimen3\font minus
  \fontdimen4\font\relax}
\providecommand{\BIBforeignlanguage}[2]{{%
\expandafter\ifx\csname l@#1\endcsname\relax
\typeout{** WARNING: IEEEtran.bst: No hyphenation pattern has been}%
\typeout{** loaded for the language `#1'. Using the pattern for}%
\typeout{** the default language instead.}%
\else
\language=\csname l@#1\endcsname
\fi
#2}}
\providecommand{\BIBdecl}{\relax}
\BIBdecl

\bibitem{rudenko2020human}
A.~Rudenko, L.~Palmieri, M.~Herman, K.~M. Kitani, D.~M. Gavrila, and K.~O.
  Arras, ``Human motion trajectory prediction: A survey,'' \emph{Int. J. Robot.
  Res.}, vol.~39, no.~8, pp. 895--935, 2020.

\bibitem{Lu2025FAPP}
M.~Lu, X.~Fan, H.~Chen, and P.~Lu, ``Fapp: Fast and adaptive perception and
  planning for uavs in dynamic cluttered environments,'' \emph{IEEE
  Transactions on Robotics}, vol.~41, pp. 871--886, 2025.

\bibitem{Kim2014Catching}
S.~Kim, A.~Shukla, and A.~Billard, ``Catching objects in flight,'' \emph{IEEE
  Transactions on Robotics}, vol.~30, no.~5, pp. 1049--1065, 2014.

\bibitem{wang2013probabilistic}
Z.~Wang, K.~M{\"u}lling, M.~P. Deisenroth, H.~Ben~Amor, D.~Vogt,
  B.~Sch{\"o}lkopf, and J.~Peters, ``Probabilistic movement modeling for
  intention inference in human--robot interaction,'' \emph{Int. J. Robot.
  Res.}, vol.~32, no.~7, pp. 841--858, 2013.

\bibitem{trick2019multimodal}
S.~Trick, D.~Koert, J.~Peters, and C.~A. Rothkopf, ``Multimodal uncertainty
  reduction for intention recognition in human-robot interaction,'' in
  \emph{2019 IEEE/RSJ IROS}.\hskip 1em plus 0.5em minus 0.4em\relax IEEE, 2019,
  pp. 7009--7016.

\bibitem{koert2019learning}
D.~Koert, J.~Pajarinen, A.~Schotschneider, S.~Trick, C.~Rothkopf, and
  J.~Peters, ``Learning intention aware online adaptation of movement
  primitives,'' \emph{IEEE RA-L}, vol.~4, no.~4, pp. 3719--3726, 2019.

\bibitem{liang2021detection}
J.~Liang, B.~I. Ahmad, M.~Jahangir, and S.~Godsill, ``Detection of malicious
  intent in non-cooperative drone surveillance,'' in \emph{2021 Sensor Signal
  Processing for Defence Conference}.\hskip 1em plus 0.5em minus 0.4em\relax
  IEEE, 2021, pp. 1--5.

\bibitem{chen2025orbital}
Y.~Chen, X.~Zhang, W.~Liao, G.~Wei, and S.~Fan, ``Orbital behavior intention
  recognition for space non-cooperative targets under multiple constraints,''
  \emph{Aerospace}, vol.~12, no.~6, p. 520, 2025.

\bibitem{alahi2016social}
A.~Alahi, K.~Goel, V.~Ramanathan, A.~Robicquet, L.~Fei-Fei, and S.~Savarese,
  ``Social lstm: Human trajectory prediction in crowded spaces,'' in
  \emph{Proceedings of the IEEE conference on computer vision and pattern
  recognition}, 2016, pp. 961--971.

\bibitem{xu2019predicting}
J.~Xu, J.~Zhao, R.~Zhou, C.~Liu, P.~Zhao, and L.~Zhao, ``Predicting
  destinations by a deep learning based approach,'' \emph{IEEE Transactions on
  Knowledge and Data Engineering}, vol.~33, no.~2, pp. 651--666, 2019.

\bibitem{gupta2018social}
A.~Gupta, J.~Johnson, L.~Fei-Fei, S.~Savarese, and A.~Alahi, ``Social gan:
  Socially acceptable trajectories with generative adversarial networks,'' in
  \emph{Proceedings of the IEEE conference on computer vision and pattern
  recognition}, 2018, pp. 2255--2264.

\bibitem{Li2024Interactive}
J.~Li, D.~Isele, K.~Lee, J.~Park, K.~Fujimura, and M.~J. Kochenderfer,
  ``Interactive autonomous navigation with internal state inference and
  interactivity estimation,'' \emph{IEEE Transactions on Robotics}, vol.~40,
  pp. 2932--2949, 2024.

\bibitem{Liao2025ASafety}
H.~Liao, Z.~Li, K.~Zhu, K.~Li, and C.~Xu, ``Sa-tp$^{2}$: A safety-aware
  trajectory prediction and planning model for autonomous driving,'' \emph{IEEE
  Transactions on Robotics}, pp. 1--20, 2025.

\bibitem{karniadakis2021physics}
G.~E. Karniadakis, I.~G. Kevrekidis, L.~Lu, P.~Perdikaris, S.~Wang, and
  L.~Yang, ``Physics-informed machine learning,'' \emph{Nature Reviews
  Physics}, vol.~3, no.~6, pp. 422--440, 2021.

\bibitem{liang2025online}
Z.~Liang, T.~Zhou, Z.~Lu, and S.~Mou, ``Online control-informed learning,''
  \emph{Transactions on Machine Learning Research}, 2025.

\bibitem{ivanovic2022expanding}
B.~Ivanovic, J.~Harrison, and M.~Pavone, ``Expanding the deployment envelope of
  behavior prediction via adaptive meta-learning,'' \emph{arXiv preprint
  arXiv:2209.11820}, 2022.

\bibitem{li2024metatra}
X.~Li, F.~Huang, Z.~Fan, F.~Mou, Y.~Hou, C.~Qian, and L.~Wen, ``Metatra:
  Meta-learning for generalized trajectory prediction in unseen domain,''
  \emph{arXiv preprint arXiv:2402.08221}, 2024.

\bibitem{yepes2007new}
J.~L. Yepes, I.~Hwang, and M.~Rotea, ``New algorithms for aircraft intent
  inference and trajectory prediction,'' \emph{Journal of Guidance, Control,
  and Dynamics}, vol.~30, no.~2, pp. 370--382, 2007.

\bibitem{agamennoni2012estimation}
G.~Agamennoni, J.~I. Nieto, and E.~M. Nebot, ``Estimation of multivehicle
  dynamics by considering contextual information,'' \emph{IEEE Transactions on
  Robotics}, vol.~28, no.~4, pp. 855--870, 2012.

\bibitem{kooij2019context}
J.~F. Kooij, F.~Flohr, E.~A. Pool, and D.~M. Gavrila, ``Context-based path
  prediction for targets with switching dynamics,'' \emph{International Journal
  of Computer Vision}, vol. 127, no.~3, pp. 239--262, 2019.

\bibitem{gan2020modeling}
R.~Gan, J.~Liang, B.~I. Ahmad, and S.~Godsill, ``Modeling intent and
  destination prediction within a bayesian framework: Predictive touch as a
  usecase,'' \emph{Data-Centric Engineering}, vol.~1, p. e12, 2020.

\bibitem{zanchettin2017probabilistic}
A.~M. Zanchettin and P.~Rocco, ``Probabilistic inference of human arm reaching
  target for effective human-robot collaboration,'' in \emph{2017 IEEE/RSJ
  IROS}.\hskip 1em plus 0.5em minus 0.4em\relax IEEE, 2017, pp. 6595--6600.

\bibitem{ahmad2015destination}
B.~I. Ahmad, J.~Murphy, P.~M. Langdon, R.~Hardy, and S.~J. Godsill,
  ``Destination inference using bridging distributions,'' in \emph{2015
  ICASSP}.\hskip 1em plus 0.5em minus 0.4em\relax IEEE, 2015, pp. 5585--5589.

\bibitem{ahmad2019bayesian}
B.~I. Ahmad, P.~M. Langdon, and S.~J. Godsill, ``A bayesian framework for
  intent prediction in object tracking,'' in \emph{2019 ICASSP}.\hskip 1em plus
  0.5em minus 0.4em\relax IEEE, 2019, pp. 8439--8443.

\bibitem{petrich2013map}
D.~Petrich, T.~Dang, D.~Kasper, G.~Breuel, and C.~Stiller, ``Map-based long
  term motion prediction for vehicles in traffic environments,'' in \emph{2013
  ITSC}.\hskip 1em plus 0.5em minus 0.4em\relax IEEE, 2013, pp. 2166--2172.

\bibitem{coscia2018long}
P.~Coscia, F.~Castaldo, F.~A. Palmieri, A.~Alahi, S.~Savarese, and L.~Ballan,
  ``Long-term path prediction in urban scenarios using circular
  distributions,'' \emph{Image and Vision Computing}, vol.~69, pp. 81--91,
  2018.

\bibitem{Kanazawa2019Adaptive}
A.~Kanazawa, J.~Kinugawa, and K.~Kosuge, ``Adaptive motion planning for a
  collaborative robot based on prediction uncertainty to enhance human safety
  and work efficiency,'' \emph{IEEE Transactions on Robotics}, vol.~35, no.~4,
  pp. 817--832, 2019.

\bibitem{Rosmann2017Online}
C.~Rösmann, M.~Oeljeklaus, F.~Hoffmann, and T.~Bertram, ``Online trajectory
  prediction and planning for social robot navigation,'' in \emph{2017 IEEE
  AIM}, 2017, pp. 1255--1260.

\bibitem{karasev2016intent}
V.~Karasev, A.~Ayvaci, B.~Heisele, and S.~Soatto, ``Intent-aware long-term
  prediction of pedestrian motion,'' in \emph{2016 IEEE ICRA}.\hskip 1em plus
  0.5em minus 0.4em\relax IEEE, 2016, pp. 2543--2549.

\bibitem{vasquez2016novel}
D.~Vasquez, ``Novel planning-based algorithms for human motion prediction,'' in
  \emph{2016 IEEE ICRA}.\hskip 1em plus 0.5em minus 0.4em\relax IEEE, 2016, pp.
  3317--3322.

\bibitem{best2015bayesian}
G.~Best and R.~Fitch, ``Bayesian intention inference for trajectory prediction
  with an unknown goal destination,'' in \emph{2015 IEEE/RSJ IROS}.\hskip 1em
  plus 0.5em minus 0.4em\relax IEEE, 2015, pp. 5817--5823.

\bibitem{yoon2021predictive}
H.~Yoon and S.~Sankaranarayanan, ``Predictive runtime monitoring for mobile
  robots using logic-based bayesian intent inference,'' in \emph{2021 IEEE
  ICRA}.\hskip 1em plus 0.5em minus 0.4em\relax IEEE, 2021, pp. 8565--8571.

\bibitem{liang2019destination}
J.~Liang, B.~I. Ahmad, R.~Gan, P.~Langdon, R.~Hardy, and S.~Godsill, ``On
  destination prediction based on markov bridging distributions,'' \emph{IEEE
  Signal Processing Letters}, vol.~26, no.~11, pp. 1663--1667, 2019.

\bibitem{schwarting2019social}
W.~Schwarting, A.~Pierson, J.~Alonso-Mora, S.~Karaman, and D.~Rus, ``Social
  behavior for autonomous vehicles,'' \emph{Proceedings of the National Academy
  of Sciences}, vol. 116, no.~50, pp. 24\,972--24\,978, 2019.

\bibitem{hu2021novel}
Y.~Hu, C.~Gao, J.~Li, W.~Jing, and Z.~Li, ``Novel trajectory prediction
  algorithms for hypersonic gliding vehicles based on maneuver mode on-line
  identification and intent inference,'' \emph{Measurement Science and
  Technology}, vol.~32, no.~11, p. 115012, 2021.

\bibitem{yang2020assisted}
X.~Yang and N.~Michael, ``Assisted mobile robot teleoperation with
  intent-aligned trajectories via biased incremental action sampling,'' in
  \emph{2020 IEEE/RSJ IROS}.\hskip 1em plus 0.5em minus 0.4em\relax IEEE, 2020,
  pp. 10\,998--11\,003.

\bibitem{devaurs2015optimal}
D.~Devaurs, T.~Sim{\'e}on, and J.~Cort{\'e}s, ``Optimal path planning in
  complex cost spaces with sampling-based algorithms,'' \emph{IEEE Trans.
  Autom. Sci. Eng.}, vol.~13, no.~2, pp. 415--424, 2015.

\bibitem{kruskal1952use}
W.~H. Kruskal and W.~A. Wallis, ``Use of ranks in one-criterion variance
  analysis,'' \emph{J. Am. Stat. Assoc.}, vol.~47, no. 260, pp. 583--621, 1952.

\bibitem{dunn1964multiple}
O.~J. Dunn, ``Multiple comparisons using rank sums,'' \emph{Technometrics},
  vol.~6, no.~3, pp. 241--252, 1964.

\bibitem{pellegrini2010improving}
S.~Pellegrini, A.~Ess, and L.~Van~Gool, ``Improving data association by joint
  modeling of pedestrian trajectories and groupings,'' in \emph{European
  conference on computer vision}.\hskip 1em plus 0.5em minus 0.4em\relax
  Springer, 2010, pp. 452--465.

\bibitem{lerner2007crowds}
A.~Lerner, Y.~Chrysanthou, and D.~Lischinski, ``Crowds by example,'' in
  \emph{Computer graphics forum}, vol.~26, no.~3.\hskip 1em plus 0.5em minus
  0.4em\relax Wiley Online Library, 2007, pp. 655--664.

\bibitem{mohamed2020social}
A.~Mohamed, K.~Qian, M.~Elhoseiny, and C.~Claudel, ``Social-stgcnn: A social
  spatio-temporal graph convolutional neural network for human trajectory
  prediction,'' in \emph{Proceedings of the IEEE/CVF conference on computer
  vision and pattern recognition}, 2020, pp. 14\,424--14\,432.

\bibitem{maeda2023fast}
T.~Maeda and N.~Ukita, ``Fast inference and update of probabilistic density
  estimation on trajectory prediction,'' in \emph{Proceedings of the IEEE/CVF
  international conference on computer vision}, 2023, pp. 9795--9805.

\bibitem{mohamed2022social}
A.~Mohamed, D.~Zhu, W.~Vu, M.~Elhoseiny, and C.~Claudel, ``Social-implicit:
  Rethinking trajectory prediction evaluation and the effectiveness of implicit
  maximum likelihood estimation,'' in \emph{European Conference on Computer
  Vision}.\hskip 1em plus 0.5em minus 0.4em\relax Springer, 2022, pp. 463--479.

\bibitem{bae2024singulartrajectory}
I.~Bae, Y.-J. Park, and H.-G. Jeon, ``Singulartrajectory: Universal trajectory
  predictor using diffusion model,'' in \emph{Proceedings of the IEEE/CVF
  Conference on Computer Vision and Pattern Recognition}, 2024, pp.
  17\,890--17\,901.

\end{thebibliography}

\begin{IEEEbiography}[{
\includegraphics[
    width=1in,
    height=1.25in,
    keepaspectratio,
    clip
]{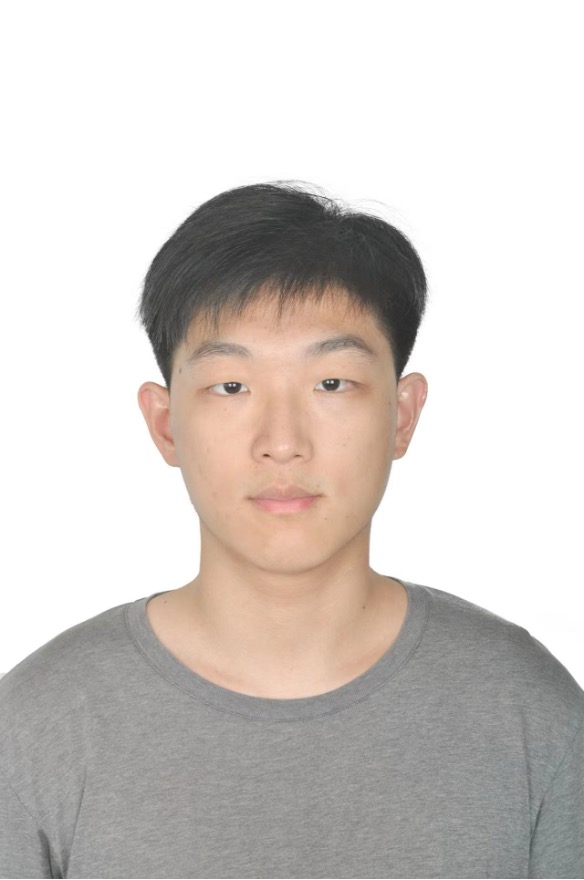}}]{Shunan Yin} received the B.E. degree in mechanics and the M.S. degree in aeronautics and astronautics from Sun Yat-sen University, Guangzhou, China. He is currently pursuing the Ph.D. degree in aeronautics and astronautics at Purdue University, West Lafayette, IN, USA. His research interests include trajectory prediction, autonomous systems, optimal control, and sampling-based planning and control.
\end{IEEEbiography}

\begin{IEEEbiography}[{\includegraphics[
  width=1in,
  height=1.25in,
  keepaspectratio
]{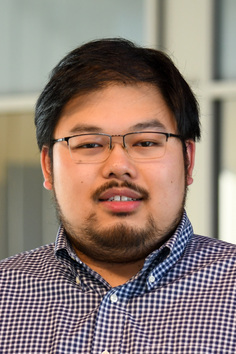}}]{Zehui Lu} received the Ph.D. degree in aeronautical and astronautical engineering from Purdue University, West Lafayette, IN, USA, in 2024. He is currently a Staff Motion Planning and Controls Engineer with Lucid Motors, Newark, CA, USA, where he develops algorithms for autonomous driving systems. Prior to this, he was a Visiting Assistant Professor with the School of Aeronautics and Astronautics, Purdue University, and held visiting research positions with Mitsubishi Electric Research Laboratories, Cambridge, MA, USA.
His research interests include real-time motion planning and control for autonomous systems, control-informed machine learning, mechatronics, and lithium-ion battery fast charging.
\end{IEEEbiography}

\begin{IEEEbiography}[{\includegraphics[
  width=1in,
  height=1.25in,
  keepaspectratio,
  clip
]{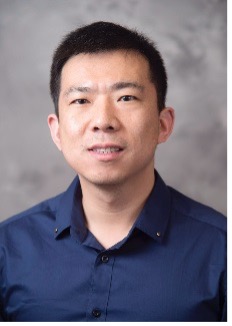}}]{Shaoshuai Mou} is the Elmer Bruhn Professor of Aeronautics and Astronautics at Purdue University. He received a Ph.D. in Electrical Engineering at Yale University in 2014, worked as a postdoc researcher at MIT for a year, and then joined Purdue University as a tenure-track assistant professor in Aug. 2015. His research has been focusing on advancing control theories with recent progress in optimization, networks and learning to address fundamental challenges in autonomous systems, with particular research interests in distributed control, optimization and learning, control of autonomous robots, human-robot teaming, cybersecurity and resilience. 
\end{IEEEbiography}

\end{document}